\documentclass[conference]{IEEEtran}
\pdfoutput=1
\usepackage{times}
\usepackage{amsfonts}
\usepackage{amsmath}
\usepackage{bm}
\usepackage{soul}

\usepackage{algorithm}
\usepackage{algorithmic}

\usepackage{authblk}
\makeatletter
\let\NAT@parse\undefined
\makeatother

\usepackage[numbers]{natbib}
\usepackage{multicol}
\usepackage{multirow}
\usepackage{graphicx}
\usepackage{color}
\usepackage[it,small]{caption}
\usepackage{subcaption}
\usepackage{dsfont}
\newcommand{\argmax}{\operatorname{arg\,max}}

\newcommand{\ve}[1]{\mathbf{#1}}

\usepackage{array}

 \belowdisplayskip \abovedisplayskip
\renewcommand{\baselinestretch}{0.97}

 \newlength\savedwidth

\newlength{\sectionReduceTop}
\newlength{\sectionReduceBot}
\newlength{\subsectionReduceTop}
\newlength{\subsectionReduceBot}
\newlength{\abstractReduceTop}
\newlength{\abstractReduceBot}
\newlength{\captionReduceTop}
\newlength{\captionReduceBot}
\newlength{\subsubsectionReduceTop}
\newlength{\subsubsectionReduceBot}

\newlength{\horSkip}
\newlength{\verSkip}

\newlength{\figureHeight}
\newcommand{\todo}[1]{\textcolor{blue}{\textbf{#1}}}

\begin{document}

\title{Brain4Cars: Car That Knows Before You Do \\ via Sensory-Fusion Deep Learning
Architecture}
\author[1,2]{Ashesh Jain}
\author[1,2]{Hema S Koppula}
\author[2]{Shane Soh}
\author[2]{Bharad Raghavan}
\author[1]{Avi Singh}
\author[3]{Ashutosh Saxena}
\affil[ ]{Cornell University$^1$, Stanford University$^2$, Brain Of Things Inc.$^3$}
\affil[ ]{{\{ashesh,hema\}@cs.cornell.edu, avisingh@iitk.ac.in, \{shanesoh,bharadr,asaxena\}@cs.stanford.edu}}

\maketitle

\IEEEpeerreviewmaketitle

\begin{abstract}

Advanced Driver Assistance Systems (ADAS) have made driving safer over the last decade. They prepare vehicles for  unsafe road conditions and alert drivers if they perform a dangerous maneuver.  However, many accidents are unavoidable because by the time drivers are alerted, it is already too late.  Anticipating maneuvers beforehand can alert drivers before they perform the  maneuver and also give ADAS more time to avoid or prepare for the danger. 

In this work we propose a vehicular sensor-rich platform and learning algorithms for  maneuver anticipation. For this purpose we equip a car with cameras, Global Positioning System (GPS), and a computing device to capture the driving context from both inside and outside of the car. In order to anticipate maneuvers, we propose a sensory-fusion deep learning architecture which jointly learns to anticipate and fuse  multiple sensory streams. Our architecture consists of Recurrent Neural Networks (RNNs) that use Long Short-Term Memory (LSTM) units to capture long temporal dependencies. We propose a novel training procedure which allows the network to predict the future given only a partial temporal context. We introduce a  diverse data set with 1180 miles of natural freeway and city driving, and show that we can anticipate  maneuvers 3.5 seconds before they occur in real-time with a precision and recall of 90.5\% and 87.4\% respectively.

\end{abstract}

\section{Introduction}
Over the last decade
cars have been equipped with various assistive technologies in order to
provide a safe driving experience. Technologies such as lane keeping, blind spot check,
pre-crash systems etc.,  are successful in alerting drivers whenever they commit a
dangerous maneuver~\citep{Laugier11}. Still in the US alone more than 33,000 people die in road
accidents every year, the majority of which are due to inappropriate
maneuvers~\cite{road_accidents}. 
We therefore need mechanisms that	 can alert drivers \textit{before} they perform a dangerous maneuver 
in order to avert many such accidents~\citep{Rueda04}.

\begin{figure}[t]
\centering
\includegraphics[width=.8\linewidth]{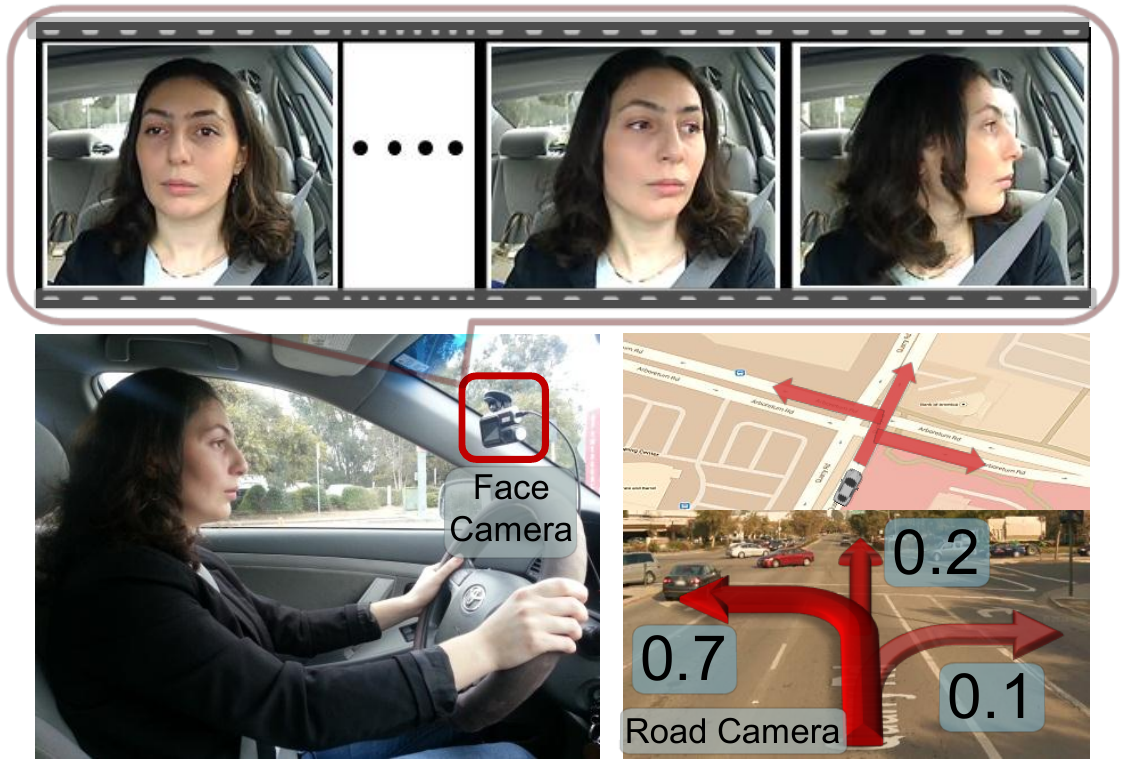}
\caption{\textbf{Anticipating maneuvers.} Our algorithm anticipates driving maneuvers performed a few seconds in the future. It uses information from multiple sources including videos, vehicle dynamics, GPS, and street maps to anticipate the probability of different future maneuvers.}
\label{fig:intro}
\end{figure}

In this work we address the problem of anticipating maneuvers that a driver is likely to perform in the next few seconds. Figure~\ref{fig:intro} shows our system anticipating a left turn maneuver a few seconds before the car reaches the intersection. Our system also outputs probabilities over the  maneuvers the driver can perform. With this prior knowledge of maneuvers, the driver assistance systems can alert drivers about possible dangers before they perform the maneuver, 
thereby giving them more time to react. Some previous works~\citep{Frohlich14,Kumar13,Morris11} also predict a driver's  future maneuver.  However, as we show in the following sections, these methods use limited context and/or do not accurately model the anticipation problem.

In order to anticipate maneuvers, we reason with the contextual information from 
the surrounding events, which we refer to as the \textit{driving context}. 
We obtain this driving context from multiple sources. We use videos
of the driver inside the car and the road in front, the vehicle's dynamics,
global position coordinates (GPS), and street maps; from this we extract a
time series of multi-modal data from both inside and outside the vehicle. The
challenge lies in modeling the temporal aspects of driving and fusing the multiple sensory streams. 
In this work we propose a specially tailored approach for anticipation in such sensory-rich settings.

Anticipation of the future actions of a human is an important perception task with applications in robotics and computer vision~\citep{Kuderer12,Ziebart09,Kitani12,Koppula13,Wang13}. It requires the prediction of future events from  a limited temporal context. This differentiates anticipation from \textit{activity recognition}~\citep{Wang13}, where the complete temporal context is available for prediction.  Furthermore, in sensory-rich robotics settings like ours, the context for anticipation comes from multiple sensors. In such scenarios the end performance of the application largely depends on how the information from different sensors are fused. Previous works on anticipation~\citep{Kitani12,Koppula13,Kuderer12} usually deal with single-data modality and do not address anticipation for sensory-rich robotics applications. Additionally, they learn representations using shallow architectures~\citep{Jain15,Kitani12,Koppula13,Kuderer12} that cannot handle long temporal dependencies~\citep{Bengio11}. 

\begin{figure}[t]
\centering	
\includegraphics[width=\linewidth]{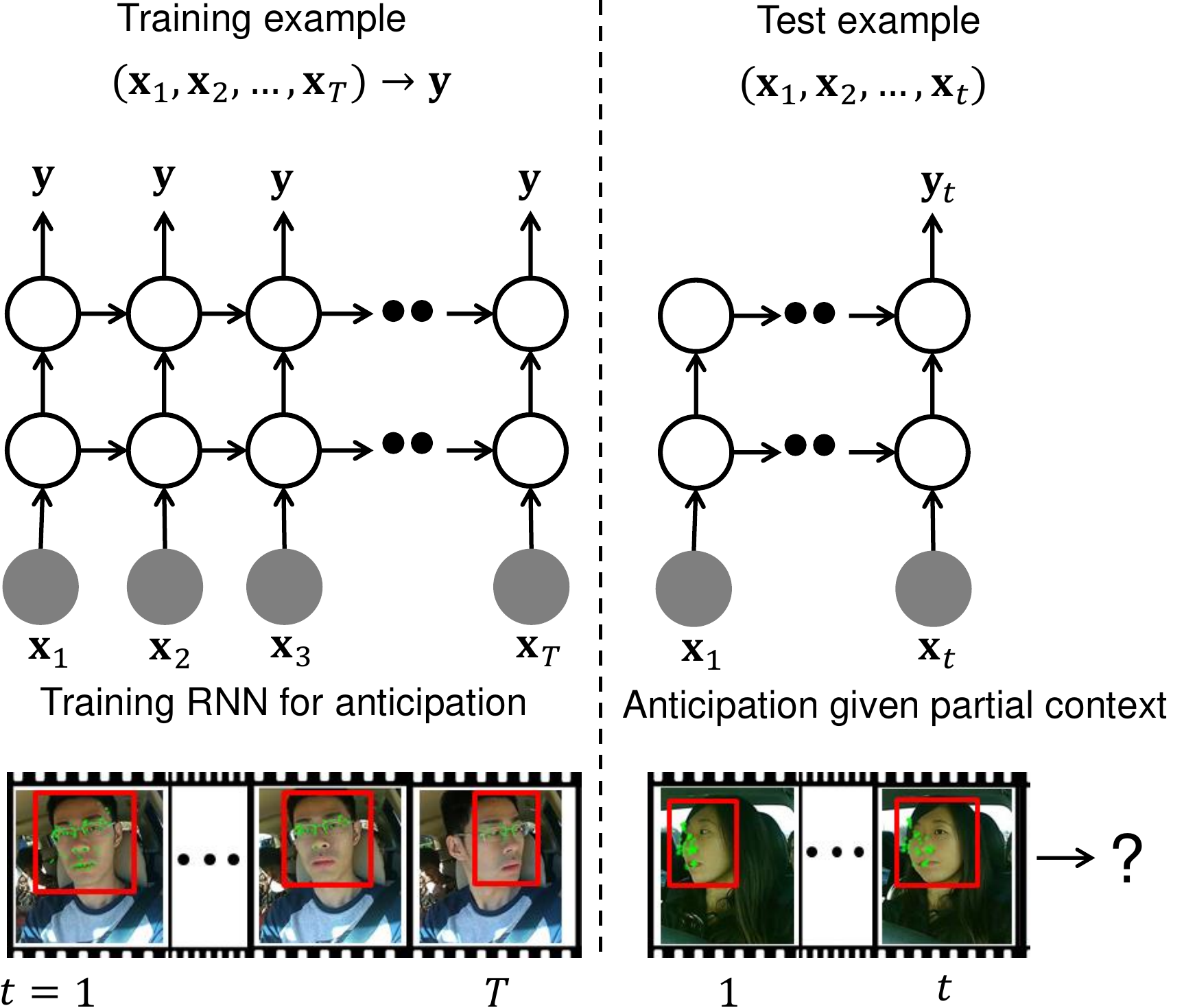}
\caption{(\textbf{Left}) Shows training RNN for anticipation in a sequence-to-sequence prediction manner. The network explicitly learns to map the partial context $(\ve{x}_1,..,\ve{x}_t)\;\forall t$ to the future event $\ve{y}$. (\textbf{Right}) At test time the network's goal is to anticipate the future event as soon as possible, i.e. by observing only a partial temporal context.}
\label{fig:introfig}
\end{figure}

In order to address the anticipation problem more generally, we propose a Recurrent Neural Network (RNN) based architecture which learns rich representations for anticipation. We focus on sensory-rich robotics applications, and our  architecture learns how to optimally fuse information from different sensors. Our approach  captures temporal dependencies by using Long Short-Term Memory (LSTM) units.  We train our architecture in a sequence-to-sequence prediction manner (Figure~\ref{fig:introfig}) such that it explicitly learns to anticipate given a partial context, and we introduce a novel loss layer which helps anticipation by preventing over-fitting. 

We evaluate our approach on a driving data set with 1180 miles of natural freeway and
city driving collected across two states -- from 10 drivers and with different kinds of driving maneuvers. The data set is challenging because of the variations in routes and traffic conditions, and the driving styles of the drivers (Figure~\ref{fig:dataset}). We demonstrate that our deep learning sensory-fusion approach anticipates 
maneuvers 3.5 seconds before they occur with {84.5\%} precision and {77.1\%} recall while using out-of-the-box face tracker. With more sophesticated 3D pose estimation of the face, our precision and recall increases to \textbf{90.5\%} and \textbf{87.4\%} respectively. We believe that our work creates scope for new ADAS features to make roads safer. In summary our key contributions are as follows:

 \begin{figure}[t]
\centering	
\includegraphics[width=.9\linewidth]{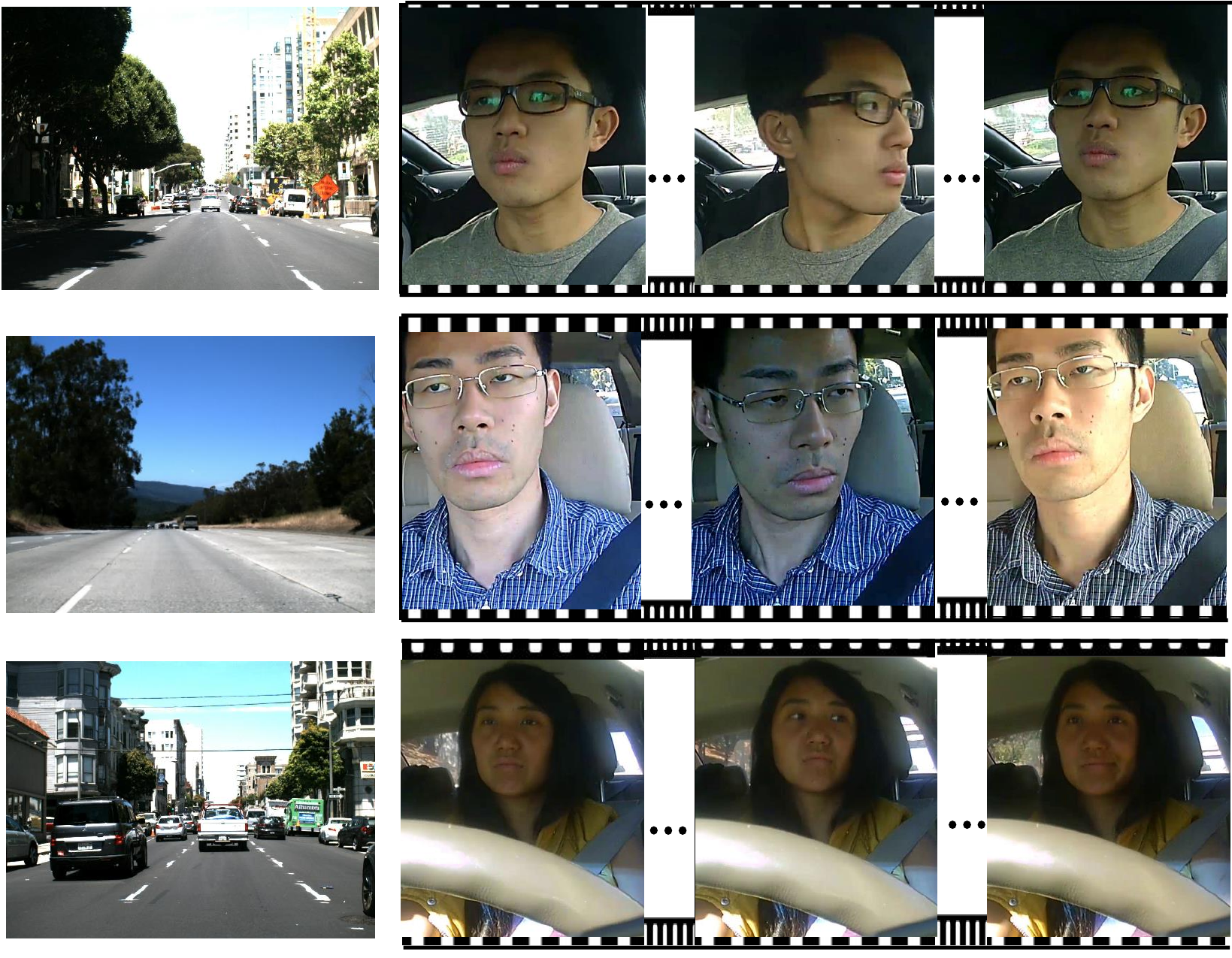}
\caption{\textbf{Variations in the data set.} {Images from the data set~\citep{Jain15} for} a left lane change. (\textbf{Left}) Views from the road facing camera. (\textbf{Right}) Driving style of the drivers vary for the same maneuver.}
\label{fig:dataset}
\end{figure}

\begin{itemize}
\itemsep0em 
\item We propose an approach for anticipating driving maneuvers several seconds in advance.
\item We propose a generic sensory-fusion RNN-LSTM architecture for anticipation in robotics applications.
\item We release the first data set of natural driving with videos from both inside and outside the car, GPS, and speed information.
\item We release an open-source deep learning package \href{https://github.com/asheshjain399/NeuralModels}{\texttt{NeuralModels}} which is especially designed for robotics applications with multiple sensory streams. 
\end{itemize}
Our data set and deep learning code are publicly available at: \url{http://www.brain4cars.com}

\section{Related Work}

Our work builds upon the previous works on assisitive vehicular technologies, anticipating human activities, learning temporal models, and computer vision methods for analyzing human face. 

\noindent \textbf{Assistive features for vehicles.} 
Latest cars available in market comes equipped with cameras and sensors to monitor the surrounding environment. Through multi-sensory fusion they provide assisitive features like lane keeping, forward collision avoidance, adaptive cruise control etc. These systems warn drivers when they perform a potentially dangerous maneuver~\citep{Shia14,Vasudevan12}. Driver monitoring for distraction and drowsiness has also been extensively researched~\citep{Fletcher05,Rezaei14}. Techniques like eye-gaze tracking are now commercially available (Seeing Machines Ltd.) and has been effective in detecting distraction. Our work complements existing ADAS and driver monitoring techniques by anticipating maneuvers several seconds before they occur.

Closely related to us are previous works on predicting the driver's intent. Vehicle trajectory has been used to predict the intent for lane change or turn maneuver~\citep{Berndt08,Frohlich14,Kumar13,Liebner12}. Most of these works ignore the rich context available from cameras, GPS, and street maps. Previous works have addressed maneuver anticipation~\citep{BoschURBAN,Morris11,Doshi11,Trivedi07} through sensory-fusion from multiple cameras, GPS, and vehicle dynamics. In particular, Morris et al.~\citep{Morris11} and Trivedi et al.~\citep{Trivedi07} used Relevance Vector Machine (RVM) for intent prediction and performed sensory fusion by concatenating feature vectors. We will show that such hand designed concatenation of features does not work well. Furthermore, these works do not model the temporal aspect of the problem properly. They assume that informative contextual cues always appear at a fixed time before the maneuver. We show that this assumption is not true, and in fact the temporal aspect of the problem should be carefully modeled. In contrast to these works, our RNN-LSTM based sensory-fusion architecture captures long temporal dependencies through its memory cell and learns rich representations for anticipation through a hierarchy of non-linear transformations of input data.  {Our work is also 
related to works on driver} behavior prediction with different sensors~\citep{Jabon10,Fletcher05,Fletcher03}, and vehicular controllers which act on these predictions~\citep{Shia14,Vasudevan12,Driggs15}.

\noindent \textbf{Anticipation and Modeling Humans.} Modeling of human motion has given rise to many applications, anticipation being one of them. Anticipating human activities has  shown to improve human-robot collaboration~\citep{Wang13,Koppula15,Mainprice13,Koppula14,Dragan12}. Similarly, forecasting human navigation trajectories has enabled robots to plan sociable trajectories around humans~\citep{Kitani12,Bennewitz05,Kuderer12,Jain15_icra}. Feature matching techniques have been proposed for anticipating human activities from videos~\citep{Ryoo11}. Modeling human preferences has enabled robots to plan good trajectories~\citep{Dragan13b,Sisbot07,Jain13b,Jain15_ijrr}. Similar to these works, we anticipate human actions, which are driving maneuvers in our case. However, the algorithms proposed in the previous works do not apply in our setting. In our case, anticipating maneuvers requires modeling the interaction between the driving context and the driver's intention. Such interactions are absent in the previous works, and they use shallow architectures~\citep{Bengio11} that do not properly model temporal aspects of human activities. They further deal with a single data modality and do not tackle the challenges of sensory-fusion. Our problem setup involves all these challenges, for which we propose a deep learning approach which efficiently handles temporal dependencies and learns to fuse multiple sensory streams.

\noindent \textbf{Analyzing the human face.} 
The vision approaches related to our work are face detection and tracking~\citep{Viola04,Zhang10}, statistical models of face~\citep{Cootes01} and pose estimation methods for face~\citep{Xiong14}. Active Appearance Model (AAM)~\citep{Cootes01} and its variants~\citep{Matthews04,Xiong13} statistically model the shape and texture of the face. AAMs have also been used to estimate the 3D-pose of a face from a single image~\citep{Xiong14} and in design of assistive features for driver monitoring~\citep{Rezaei14,Tawari14b}. In our approach we adapt off-the-shelf available face detection~\citep{Viola04} and tracking algorithms~\citep{Shi94} (see Section~\ref{sec:features}). Our approach allows us to easily experiment with more advanced face detection and tracking algorithms. We demonstrate this by using the Constrained Local Neural Field (CLNF) model~\citep{Baltrusaitis13} and tracking 68 fixed landmark points on the driver's face and estimating the 3D head-pose.

\noindent \textbf{Learning temporal models.} Temporal models are commonly used to model human activities~\citep{Koppula13c,Morency07,Wang06,Wang05}. These models have been used in both discriminative and generative fashions. The discriminative temporal models are mostly inspired by the Conditional Random Field (CRF)~\citep{Lafferty01} which  captures the temporal structure of the problem. Wang et al.~\citep{Wang05} and Morency et al.~\citep{Morency07} propose dynamic extensions of the CRF for image segmentation and gesture recognition respectively. On the other hand, generative approaches for temporal modeling include various filtering methods, such as Kalman and particle filters~\citep{Thrun05}, Hidden Markov Models, and many types of Dynamic Bayesian Networks~\citep{Murphy12}. Some previous works~\citep{Berndt08,Kuge00,Oliver00} used HMMs to model different aspects of the driver's behaviour. Most of these generative approaches model how latent (hidden) states influence the observations. However, in our problem both the latent states and the observations influence each other. 
In the following sections, we will describe the Autoregressive Input-Output HMM (AIO-HMM) for maneuver anticipation~\citep{Jain15} and will use it as a baseline to compare our deep learning approach. Unlike AIO-HMM our deep architecture have internal memory which allows it to handle long temporal dependencies~\citep{Hihi95}. Furthermore, the input features undergo a hierarchy of  non-linear transformation through the deep architecture which allows learning rich representations. 

Two building blocks of our architecture are Recurrent Neural Networks (RNNs)~\citep{Pascanu12} and Long Short-Term Memory (LSTM) units~\citep{Hochreiter97}. Our work draws upon ideas from previous works on RNNs and LSTM from the language~\citep{Sutskever14}, speech~\citep{Hannun14}, and vision~\citep{Donahue15} communities. 
Our approach to the joint training of multiple RNNs is related to the recent work on hierarchical RNNs~\citep{Du15}. We consider RNNs in multi-modal setting, which is related to the recent use of RNNs in image-captioning~\citep{Donahue15}. Our contribution lies in formulating activity anticipation in a deep learning framework using RNNs with LSTM units. We focus on sensory-rich robotics applications, and our architecture extends previous works doing sensory-fusion with feed-forward networks~\citep{Ngiam11,Sung15} to the fusion of temporal streams. Using our architecture we demonstrate state-of-the-art on maneuver anticipation.

\begin{figure*}[t]
\centering
\includegraphics[width=\linewidth]{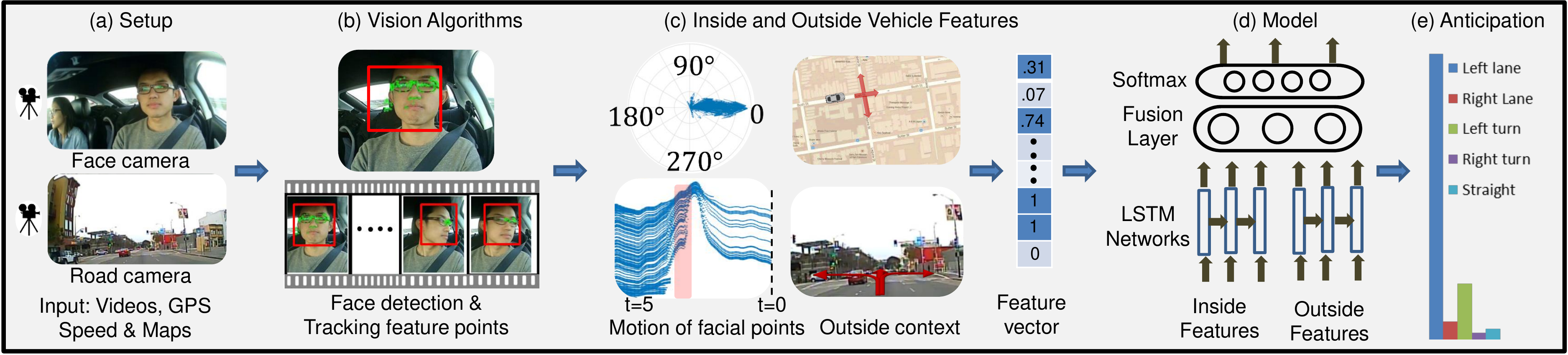}
\caption{\textbf{System Overview.} Our system anticipating a left lane
change maneuver. (a) We process multi-modal data including 
GPS, speed, street maps, and events inside and
outside of the vehicle using video cameras. (b) Vision pipeline extracts visual cues such as driver's head movements.
 (c) The inside and outside driving context is processed to extract expressive
 features. (d,e) Using our deep learning architecture we fuse the information from outside and inside the vehicle and anticipate the probability of each maneuver.}
\label{fig:system}
\end{figure*}

\section{Overview}
\label{sec:overview}

We first give an overview of the maneuver anticipation problem and then describe our system. 
\subsection{Problem Overview}
Our goal is to anticipate  driving maneuvers a few seconds before they occur. This includes anticipating a lane change before the wheels touch the lane markings or anticipating if the driver keeps straight or makes a turn when approaching an intersection. This is a challenging problem for multiple reasons. First, it requires the modeling of context from different sources. Information from a single source, such as a camera  capturing events outside the car, is not sufficiently rich.  Additional visual information from within the car can also be used.  For example, the driver's head movements are useful for anticipation -- drivers typically check for the side traffic while changing lanes and scan the cross traffic at intersections.

\begin{figure}[t]
\centering
\includegraphics[width=.85\linewidth]{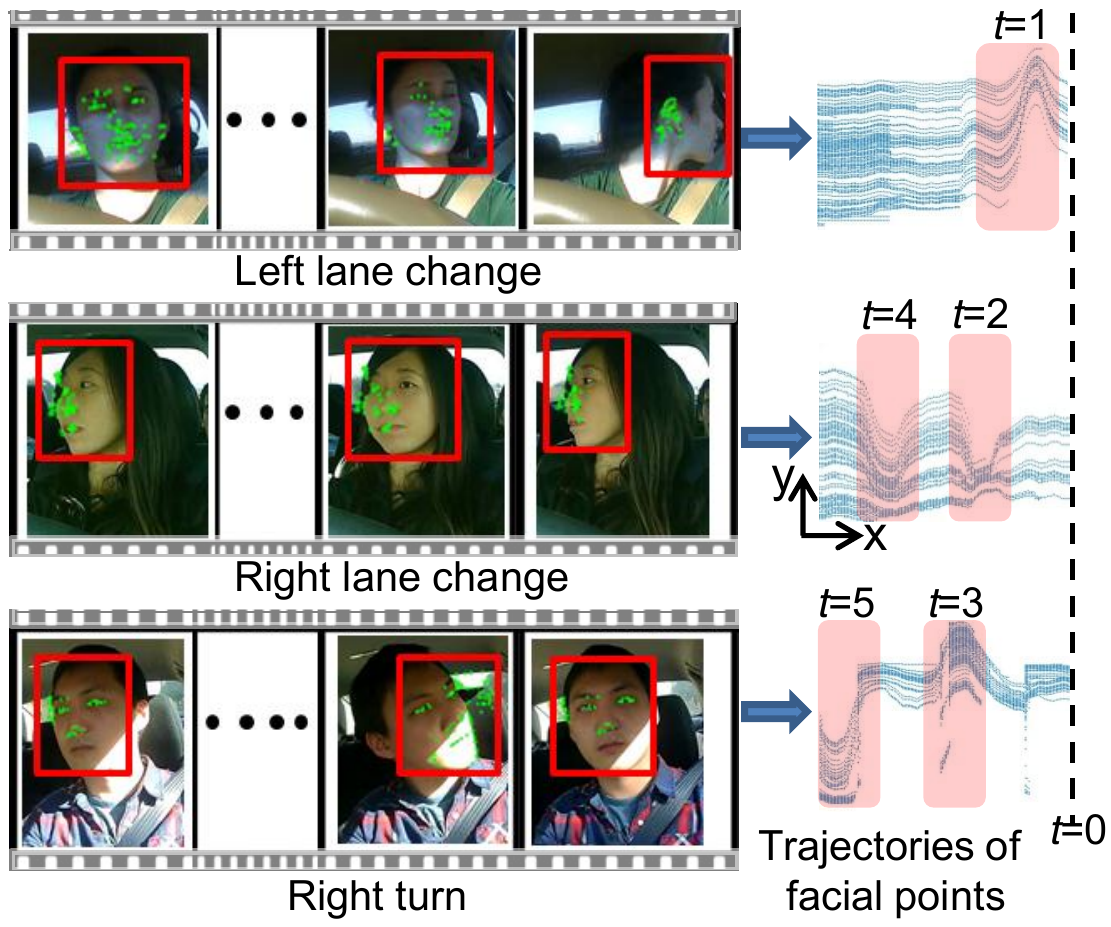}
\caption{\textbf{Variable time occurrence of events.} \textit{Left}: The events inside the vehicle before the maneuvers. We track the driver's face along with many facial points. \textit{Right}: The trajectories generated by the horizontal motion of facial points (pixels) `t' seconds before the maneuver. X-axis is the time and Y-axis is the pixels' horizontal coordinates. Informative cues appear during the shaded time interval. Such cues occur at variable times before the maneuver, and the order in which the cues appear is also important.}
\label{fig:time_variance}
\end{figure}

Second, reasoning about maneuvers should take into account the driving context at both local and global levels. Local context requires modeling events in vehicle's vicinity such as the surrounding vision, GPS, and speed information. On the other hand, factors that influence the overall route contributes to the global context, such as the driver's final destination. Third, the informative cues necessary for anticipation appear at variable times before the maneuver, as illustrated in Figure~\ref{fig:time_variance}. In particular, the time interval between the driver's head movement and the occurrence of the maneuver depends on many factors such as the speed, traffic conditions, etc.

In addition, appropriately fusing the information from multiple sensors is crucial for anticipation. Simple sensory fusion approaches like concatenation of feature vectors performs poorly, as we demonstrate through experiments. In our proposed approach we learn a neural network layer for fusing the temporal streams of data coming from different sensors. Our resulting architecture is end-to-end trainable via back propagation, and we jointly train it to: (i) model the temporal aspects of the problem; (ii) fuse multiple sensory streams; and (iii) anticipate maneuvers.

\subsection{System Overview}

For maneuver anticipation our vehicular sensory platform includes the following (as shown in Figure~\ref{fig:system}): 
\begin{enumerate}
\item A driver-facing camera inside the vehicle. We mount this camera on the dashboard and use it to track the driver's head movements. This camera operates at 25 fps.
\item A camera facing the road is mounted on the dashboard to capture the (outside) view in front of the car. This camera operates at 30 fps. The video from this camera enables additional reasoning on maneuvers. For example, when the vehicle is in the left-most lane, the only safe maneuvers are a right-lane change or keeping straight, unless the vehicle is approaching an intersection.
\item A speed logger for vehicle dynamics because maneuvers correlate with the vehicle's speed, e.g., turns usually happen at  lower speeds than lane changes.
\item A Global Positioning System (GPS) for localizing the vehicle on the map. This enables us to detect upcoming road artifacts such as intersections, highway exits, etc.
\end{enumerate}

Using this system we collect 1180 miles of natural city and freeway driving data from 10 drivers. We denote the information from sensors with feature vector $\ve{x}$. Our vehicular systems gives a temporal sequence of feature vectors $\{(\ve{x}_1,\ve{x}_2,...,\ve{x}_t,...)\}$.  For now we do not distinguish between the information from different sensors, later in Section~\ref{sec:sensor-fusion-rnn} we introduce sensory fusion. In Section~\ref{sec:features} we formally define our feature representations and describe our data set in Section~\ref{subsec:dataset}. 
We now formally define anticipation and present our deep learning architecture.

\section{Preliminaries}

We now formally define anticipation and then present our Recurrent Neural
Network architecture. The goal of anticipation is to predict an event several
seconds before it happens given the contextual information up to the present
time. The future event can be one of multiple possibilities. At training
time a set of temporal sequences of {observations} and events
$\{(\mathbf{x}_1,\mathbf{x}_2,...,\mathbf{x}_T)_j,\ve{y}_j\}_{j=1}^N$ is provided
where $\mathbf{x}_t$ is the {observation} at time $t$, $\ve{y}$ is the representation of the event (described below) that happens at the end of the sequence  at $t= T$,
and $j$ is the sequence index. At test
time, however, the algorithm receives an {observation} $\mathbf{x}_t$ at each time step,
and its goal is to predict the future event as early as possible, i.e. by
observing only a partial sequence of {observations} $\{(\mathbf{x}_1,...,\mathbf{x}_t)
| t < T\}$. This differentiates anticipation from \textit{activity recognition}~\citep{Wang13b,Koppula13b} where in the latter  the complete {observation} sequence is available at  test time. In this paper, $\ve{x}_t$ is a real-valued feature vector and $\ve{y} =[y^1,...,y^K]$ is a vector of size $K$ (the number of events), where $y^k$ denotes the probability of the temporal sequence belonging to event the $k$ such that $\sum_{k=1}^K y^k = 1$. At the time of training, $\ve{y}$ takes the form of a one-hot vector with the entry in $\ve{y}$ corresponding to the ground truth event as $1$ and the rest $0$.

In this work we propose a deep RNN architecture with Long Short-Term Memory
(LSTM) units~\citep{Hochreiter97} for anticipation. Below we give an overview of
the standard RNN and LSTM which form the building blocks of our architecture.

\subsection{Recurrent Neural Networks}

A standard RNN~\citep{Pascanu12} takes in a temporal sequence of vectors $(\mathbf{x}_1,\mathbf{x}_2,...,\mathbf{x}_T)$ as input, and outputs a sequence of vectors $(\mathbf{h}_1,\mathbf{h}_2,...,\mathbf{h}_T)$ also known as high-level representations. The representations are generated by non-linear transformation of the input sequence from $t=1	$ to $T$, as described in the equations below.
\begin{align}
\label{eq:h-rnn} \mathbf{h}_t &= f(\mathbf{W}\mathbf{x}_t + \mathbf{H}\mathbf{h}_{t-1} + \mathbf{b})\\
\label{eq:y-rnn} \mathbf{y}_t &= \texttt{softmax} (\mathbf{W}_y \mathbf{h}_t + \mathbf{b}_y)
\end{align}  
where $f$ is a non-linear function applied element-wise, and $\mathbf{y}_t$ is the \texttt{softmax} probabilities of the events having seen the {observations} up to $\mathbf{x}_t$.  $\mathbf{W}$, $\mathbf{H}$,  $\mathbf{b}$, $\mathbf{W}_y$,  $\mathbf{b}_y$ are the parameters that are learned. Matrices are denoted with bold, capital letters, and vectors are denoted with bold, lower-case letters.  In a standard RNN a common choice for $f$ is \texttt{tanh} or \texttt{sigmoid}. RNNs with this choice of $f$ suffer from a well-studied problem of \textit{vanishing gradients}~\citep{Pascanu12}, and hence are poor at capturing long temporal dependencies which are essential for anticipation. A common remedy to vanishing gradients is to replace \texttt{tanh} non-linearities by Long Short-Term Memory cells~\citep{Hochreiter97}.  We now give an overview of LSTM and then describe our model for anticipation. 

\subsection{Long-Short Term Memory Cells}

LSTM is a network of neurons that implements a memory cell~\citep{Hochreiter97}. The central idea behind LSTM is that the memory cell can maintain its state over time. When combined with RNN, LSTM units allow the recurrent network to remember long term context dependencies.

\begin{figure}[t]
\centering	
\includegraphics[width=.8\linewidth]{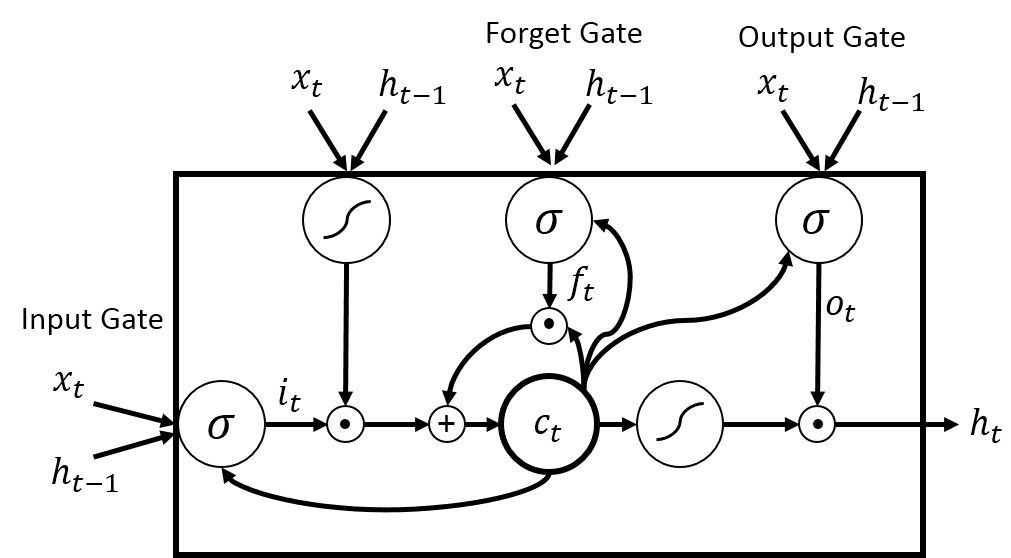}
\caption{\textbf{ Internal working of an LSTM unit.}}
\label{fig:lstm}
\end{figure}

LSTM consists of three gates -- input gate $\mathbf{i}$, output gate $\mathbf{o}$, and forget gate $\mathbf{f}$ -- and a memory cell $\mathbf{c}$. {See Figure~\ref{fig:lstm} for an illustration.} 
At each time step $t$, LSTM first computes its gates' activations \{$\mathbf{i}_t$,$\mathbf{f}_t$\}~\eqref{eq:i-lstm}\eqref{eq:f-lstm} and updates its memory cell from $\mathbf{c}_{t-1}$ to $\mathbf{c}_t$~\eqref{eq:c-lstm}, it then computes the output gate activation $\mathbf{o}_t$~\eqref{eq:o-lstm}, and finally outputs a hidden representation $\mathbf{h}_t$~\eqref{eq:h-lstm}. The inputs into LSTM are the {observations} $\mathbf{x}_t$ and the hidden representation from the previous time step $\mathbf{h}_{t-1}$. LSTM applies the following set of update operations:  
\begin{align}
\label{eq:i-lstm} \ve{i}_t &= \sigma(\ve{W}_{i}\ve{x}_t + \ve{U}_i \ve{h}_{t-1} + \ve{V}_i \ve{c}_{t-1} + \ve{b}_i)\\
\label{eq:f-lstm} \ve{f}_t &= \sigma(\ve{W}_{f}\ve{x}_t + \ve{U}_f \ve{h}_{t-1} + \ve{V}_f \ve{c}_{t-1} +\ve{b}_f)\\
\label{eq:c-lstm} \ve{c}_t &= \ve{f}_t \odot \ve{c}_{t-1} + \ve{i}_t \odot	 \texttt{tanh} (\ve{W}_{c}\ve{x}_t + \ve{U}_c \ve{h}_{t-1} +\ve{b}_c)\\
\label{eq:o-lstm} \ve{o}_t &= \sigma(\ve{W}_{o}\ve{x}_t + \ve{U}_o \ve{h}_{t-1} + \ve{V}_o \ve{c}_{t} +\ve{b}_o)\\
\label{eq:h-lstm} \ve{h}_t &= \ve{o}_t \odot \texttt{tanh}(\ve{c}_{t}) 
\end{align}

where $\odot$ is an element-wise product and $\sigma$ is the logistic function. $\sigma$ and \texttt{tanh} are applied element-wise. $\ve{W}_*$, $\ve{V}_*$, $\ve{U}_*$, and $\ve{b}_*$ are the parameters, further the weight matrices $\ve{V}_*$ are diagonal. The input and forget gates of LSTM participate in updating the memory cell~\eqref{eq:c-lstm}. More specifically, forget gate controls the part of memory to forget, and the input gate computes new values based on the current {observation} that are written to the memory cell. The output gate together with the memory cell computes the hidden representation~\eqref{eq:h-lstm}. Since  LSTM cell activation involves \textit{summation} over time~\eqref{eq:c-lstm} and derivatives distribute over sums, the gradient in LSTM gets propagated over a longer time before vanishing. In the standard RNN, we replace the non-linear $f$ in equation~\eqref{eq:h-rnn} by the LSTM equations given above in order to capture long temporal dependencies. We use the following shorthand notation to denote the recurrent LSTM operation.
\begin{equation}
(\ve{h}_t,\ve{c}_t) = \text{LSTM}(\ve{x}_t,\ve{h}_{t-1},\ve{c}_{t-1})
\end{equation}

We now describe our RNN architecture with LSTM units for anticipation. Following which we will describe a particular instantiation of our architecture for maneuver anticipation where the {observations} $\ve{x}$ come from multiple sources.

\section{Network Architecture for Anticipation}
\label{sec:network}
In order to anticipate, an algorithm must learn to predict the future {given only} a partial temporal context. This makes anticipation challenging and also differentiates it from activity recognition. Previous works treat anticipation as a recognition problem~\citep{Koppula13,Morris11,Ryoo11} and train discriminative classifiers (such as SVM or CRF) on the complete temporal context. However, at test time {these} classifiers only observe  a partial temporal context and make predictions within a filtering framework. We model anticipation with a recurrent architecture which unfolds through time. This lets us train a single classifier that learns to handle partial temporal context of varying lengths.

Furthermore, anticipation in robotics applications is challenging because the contextual information can come from multiple sensors with different data modalities. Examples include autonomous vehicles that reason from multiple sensors~\citep{Geiger12} or robots that jointly reason over perception and language instructions~\citep{Misra14}. In such applications the way information from different sensors is fused is critical to the application's final performance. We therefore build an end-to-end deep learning architecture which jointly learns to anticipate and fuse information from different sensors.   

\subsection{RNN with LSTM units for anticipation}

At the time of training, we observe the complete temporal {observation sequence} and the event  $\{(\mathbf{x}_1,\mathbf{x}_2,...,\mathbf{x}_T),\ve{y}\}$. Our goal is to train a network which predicts the future event given a partial temporal {observation sequence} $\{(\mathbf{x}_1,\mathbf{x}_2,...,\mathbf{x}_t) 	| t < T\}$. We do so by training an RNN in a sequence-to-sequence prediction manner. Given training examples $\{(\mathbf{x}_1,\mathbf{x}_2,...,\mathbf{x}_T)_j,\ve{y}_j\}_{j=1}^N$ we train an RNN with LSTM units to map the sequence of {observations} $(\mathbf{x}_1,\mathbf{x}_2,...,\mathbf{x}_T)$ to the sequence of events  $(\ve{y}_1,...,\ve{y}_T)$ such that $\ve{y}_{t} = \ve{y}, \forall t$, as shown in Fig.~\ref{fig:introfig}. Trained in this manner, our RNN will attempt to map all sequences of partial {observations} $(\mathbf{x}_1,\mathbf{x}_2,...,\mathbf{x}_t)$~$\forall t \leq~T$ to the future event $\ve{y}$. This way our model explicitly learns to anticipate. We additionally use LSTM units which prevents the gradients from vanishing and allows our model to capture long temporal dependencies in human activities.\footnote{Driving maneuvers can take up to 6 seconds and the value of T can go up to 150 with a camera frame rate of 25 fps.}


\subsection{Fusion-RNN: Sensory fusion RNN for anticipation}
\label{sec:sensor-fusion-rnn}

We now present an instantiation of our RNN architecture for fusing two sensory streams: $\{(\ve{x}_1,...,\ve{x}_T), \;(\ve{z}_1,...,\ve{z}_T)\}$. In the next section we will describe these streams for maneuver anticipation.

\begin{figure}[t]
\centering	
\includegraphics[width=.9\linewidth]{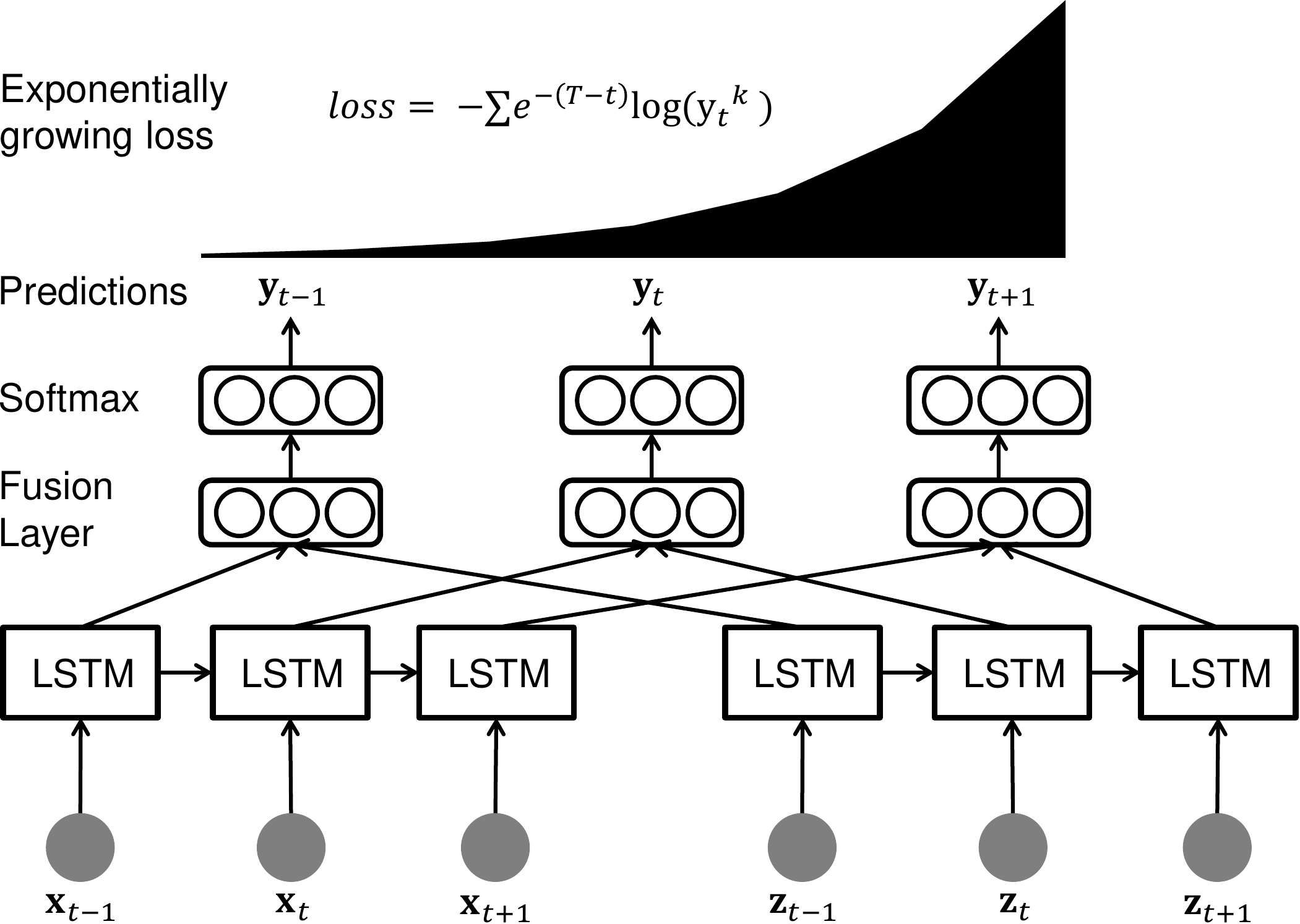}
\caption{\textbf{Sensory fusion RNN for anticipation.} (\textbf{Bottom}) In the Fusion-RNN each sensory stream is passed through their independent RNN. (\textbf{Middle}) High-level representations from RNNs are then combined through a fusion layer. (\textbf{Top}) In order to prevent over-fitting early in time the loss exponentially increases with time.}
\label{fig:fusion}
\end{figure}

An obvious way to allow sensory fusion in the RNN is by concatenating the streams, i.e. using $([\ve{x}_1;\ve{z}_1],...,[\ve{x}_T;\ve{z}_T])$ as input to the RNN. However, we found that this sort of simple concatenation performs poorly. We instead learn a sensory fusion layer which combines the high-level representations of sensor data. Our proposed architecture first passes the two sensory streams $\{(\ve{x}_1,...,\ve{x}_T), \;(\ve{z}_1,...,\ve{z}_T)\}$ independently through separate  RNNs~\eqref{eq:f-rnn1} and~\eqref{eq:f-rnn2}.  
The high level representations from both RNNs $\{(\ve{h}_1^x,...,\ve{h}_T^x), \;(\ve{h}_1^z,...,\ve{h}_T^z)$ are then concatenated at each time step $t$ and passed through a fully connected (fusion) layer which fuses the two representations~\eqref{eq:f-fusion}, {as shown in Figure~\ref{fig:fusion}}. The output representation from the fusion layer is then passed to the softmax layer for anticipation~\eqref{eq:f-output}. The following operations are performed from $t=1$ to $T$.
\begin{align}
\label{eq:f-rnn1} (\ve{h}_t^x,\ve{c}_t^x) &= \text{LSTM}_x(\ve{x}_t,\ve{h}_{t-1}^x,\ve{c}_{t-1}^x)\\
\label{eq:f-rnn2} (\ve{h}_t^z,\ve{c}_t^z) &= \text{LSTM}_z(\ve{z}_t,\ve{h}_{t-1}^z,\ve{c}_{t-1}^z)\\
\label{eq:f-fusion} \text{Sensory fusion:  } \ve{e}_t &= \texttt{tanh}(\ve{W}_f[\ve{h}_t^x;\ve{h}_t^z] + \ve{b}_f)\\
\label{eq:f-output} \ve{y}_t &= \texttt{softmax}(\ve{W}_y\ve{e}_t + \ve{b}_y)
\end{align}	 
where $\ve{W}_*$ and $\ve{b}_*$ are model parameters, and $\text{LSTM}_x$ and $\text{LSTM}_z$ process the sensory streams $(\ve{x}_1,...,\ve{x}_T)$ and $(\ve{z}_1,...,\ve{z}_T)$ respectively. {The same framework can be extended to handle more sensory streams.} 

\subsection{Exponential loss-layer for anticipation.} 
We propose a new loss layer which encourages the architecture to anticipate early while also ensuring that the architecture does not over-fit the training data early enough in time when there is not enough context for anticipation. 
When using the standard softmax loss, the architecture suffers a loss of $-\log(y_t^k)$ for the mistakes it makes at each time step, where $y_t^k$ is the probability of the ground truth event $k$ computed by the architecture using Eq.~\eqref{eq:f-output}. We propose to modify this loss by multiplying it with an exponential term as illustrated in Figure~\ref{fig:fusion}. Under this new scheme, the loss exponentially grows with time as shown below. 
\begin{align}
\label{eq:loss}
loss = \sum_{j=1}^N \sum_{t=1}^T  - e^{-(T-t)}\log (y_t^k)
\end{align}
This loss penalizes the RNN exponentially more for the mistakes it makes as it sees more {observations}. This encourages the model to fix mistakes as early as it can in time. The loss in equation~\ref{eq:loss} also penalizes the network less on mistakes made early in time when there is not enough context available. This way it acts like a regularizer and reduces the risk to over-fit very early in time. 

\section{Features}
\label{sec:features}

\begin{figure}[t]
\centering
\includegraphics[width=.9\linewidth]{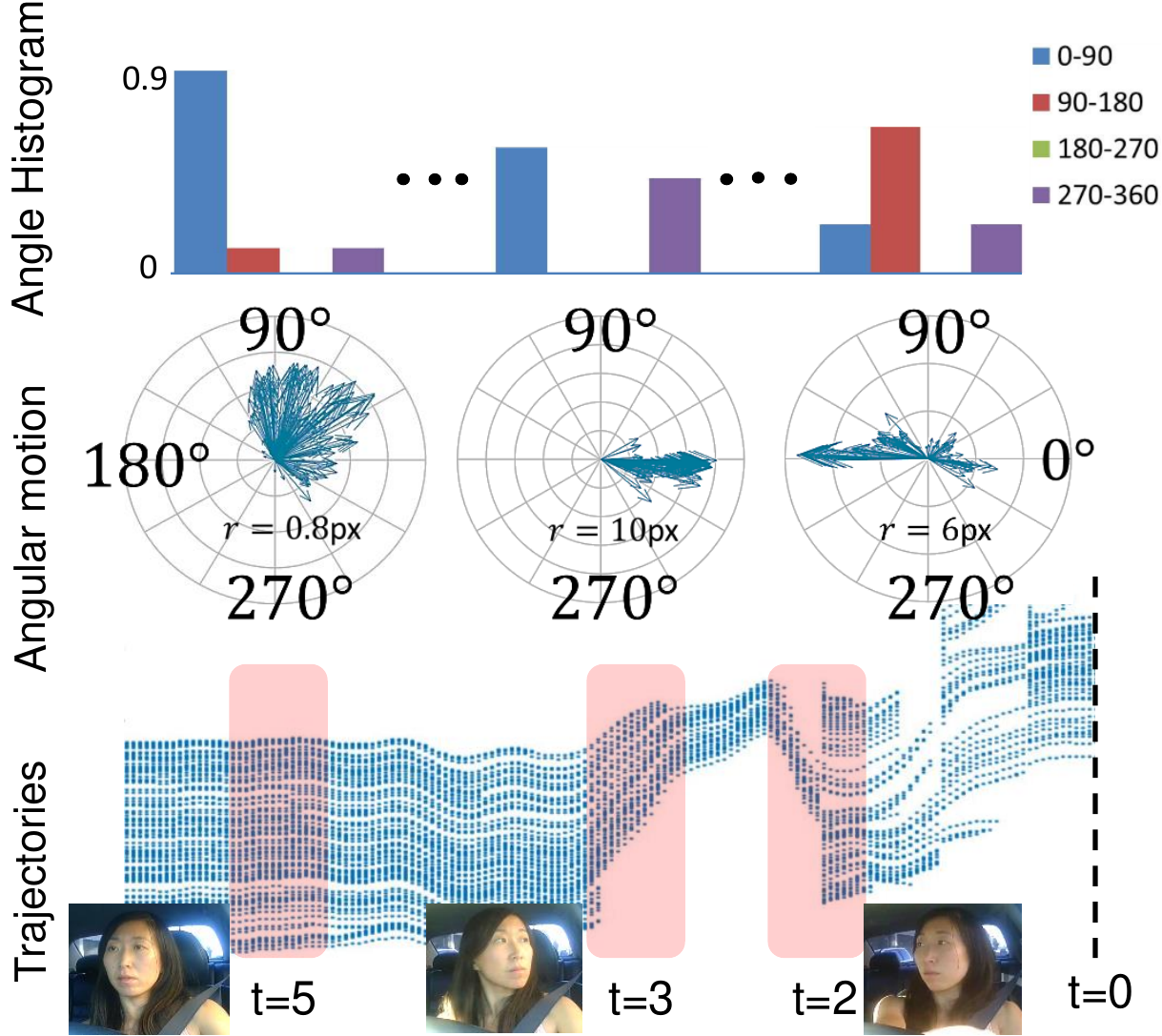}
\caption{\textbf{Inside vehicle feature extraction.} The angular histogram features extracted at three different time steps for a left turn maneuver. \textit{Bottom}: Trajectories for the horizontal motion of tracked facial pixels `t' seconds before the maneuver. At t=5 seconds before the maneuver the driver is looking straight, at t=3 looks (left) in the direction of maneuver, and at t=2 looks (right) in opposite direction for the crossing traffic. \textit{Middle}: Average motion vector of tracked facial pixels in polar coordinates. $r$ is the average movement of pixels and arrow indicates the direction in which the face moves when looking from the camera. \textit{Top}: Normalized angular histogram features. }
\label{fig:face_feature}
\end{figure}

We extract features by processing the inside and outside driving contexts. We do this by  grouping the overall contextual information from the sensors into: (i) the context from inside the vehicle, which comes from the driver facing camera and is represented as temporal sequence of features $(\ve{z}_1,...,\ve{z}_T)$; and (ii) the context from outside the vehicle, which comes from the remaining sensors: GPS, road facing camera, and street maps. We represent the outside context with $(\ve{x}_1,...,\ve{x}_T)$. In order to anticipate maneuvers, our RNN architecture (Figure~\ref{fig:fusion}) processes the temporal context $\{(\ve{x}_1,...,\ve{x}_t),\; (\ve{z}_1,...,\ve{z}_t)\}$  at every time step $t$, and outputs softmax probabilities $\ve{y}_t$ for the following five maneuvers:  $\mathcal{M}=$~\{\textit{left turn}, \textit{right turn}, \textit{left lane change}, \textit{right lane change}, \textit{straight driving}\}.

\subsection{Inside-vehicle features.}\label{subsec:inside_features} 
The inside features $\ve{z}_t$ capture the driver's head movements at each time instant $t$.  Our vision pipeline consists of face detection, tracking, and feature extraction modules. We extract head motion features per-frame,  denoted by  $\phi(\text{face})$. We compute $\ve{z}_t$ by aggregating  $\phi(\text{face})$ for every 20 frames, i.e., $\ve{z}_t = \sum_{i=1}^{20}\phi(\text{face}_i)/\|\sum_{i=1}^{20}\phi(\text{face}_i)\|$. 

\noindent \textit{Face detection and tracking.} We detect
the driver's face using a trained Viola-Jones face detector~\citep{Viola04}. From the detected face, we first extract visually discriminative (facial) points using the Shi-Tomasi corner
detector \citep{Shi94} and then track those facial points using the Kanade-Lucas-Tomasi (KLT) tracker \citep{Lucas81,Shi94,Tomasi91}. However, the tracking may accumulate errors over time because of changes in illumination due to the shadows of trees,  traffic, etc. We therefore constrain the tracked facial points to follow a projective transformation and remove the incorrectly tracked points using the RANSAC algorithm. While tracking the facial points, we  lose some of the tracked  points with every new frame. To address this problem, we  re-initialize the tracker with new discriminative facial points once the number of tracked  points falls below a threshold \citep{Kalal10}. 
\vspace{0.05in}

\noindent \textit{Head motion features.} 
For maneuver anticipation the horizontal movement of the face and its angular rotation (\textit{yaw}) are particularly important.  From the face tracking we obtain \textit{face tracks}, which are 2D trajectories of the tracked facial points in the image plane. Figure~\ref{fig:face_feature} (bottom) shows how the horizontal coordinates of the tracked facial points vary with time before a left turn maneuver. We represent the driver's face movements and rotations with histogram features. In particular, we take matching facial points between  successive frames and create histograms of their corresponding horizontal motions (in pixels) and angular motions in the image plane (Figure~\ref{fig:face_feature}). We bin the horizontal and angular motions using $[\leq-2,\;-2\;\text{to}\;0,\;0\;\text{to}\;2,\;\geq2]$ and  $[0\;\text{to}\;\frac{\pi}{2},\;\frac{\pi}{2}\;\text{to}\;\pi,\;\pi\;\text{to}\;\frac{3\pi}{2},\;\frac{3\pi}{2}\;\text{to}\;2\pi]$,
respectively. We also calculate the mean movement of the driver's face center. This gives us $\phi(\text{face})\in\mathbb{R}^9$ facial features per-frame. The driver's eye-gaze is also useful a feature. However, robustly estimating 3D eye-gaze in outside environment is still a topic of research, and orthogonal to this work on anticipation. We therefore do not consider eye-gaze features.  

\begin{figure}[t]
\centering
\includegraphics[width=.9\linewidth]{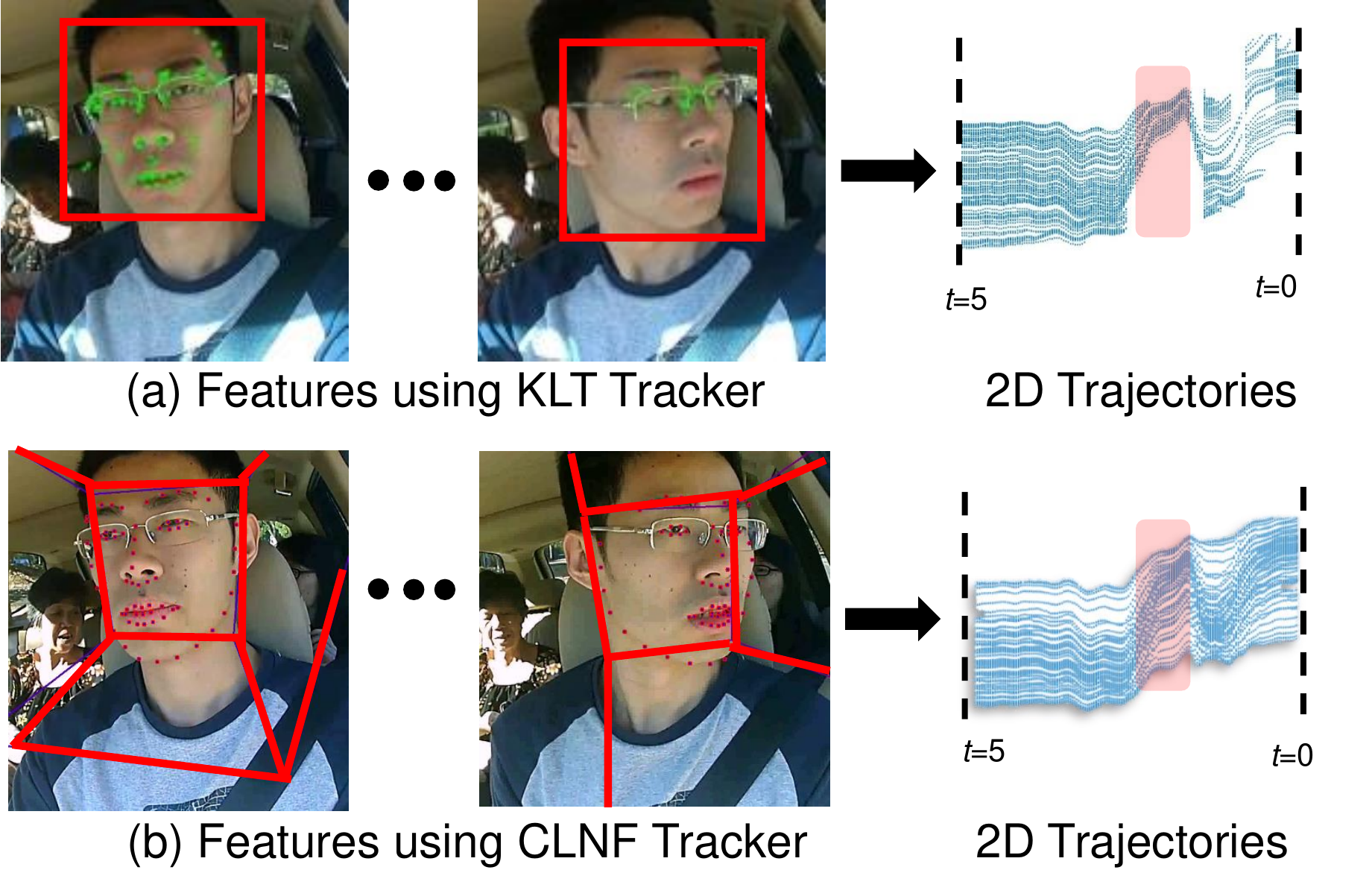}
\caption{\textbf{Improved features for maneuver anticipation.} We track facial landmark points using the CLNF tracker~\citep{Baltrusaitis13} which results in more consistent 2D trajectories as compared to the KLT tracker~\citep{Shi94} used by Jain et al.~\citep{Jain15}. Furthermore, the CLNF also gives an estimate of the driver's 3D head pose.}
\label{fig:feature_compare}
\end{figure}

\noindent \textit{3D head pose and facial landmark features.}
Our framework is flexible and allows incorporating more advanced face detection and tracking algorithms. For example we replace the KLT tracker described above with  the Constrained Local Neural Field (CLNF) model~\citep{Baltrusaitis13} and track 68 fixed  landmark points on the driver's face. CLNF is particularly well suited for driving scenarios due its ability to handle a wide range of head pose and illumination variations. As shown in Figure~\ref{fig:feature_compare}, CLNF offers us two distinct benefits over the features from KLT (i) while discriminative facial points may change from situation to situation, tracking fixed landmarks results in consistent optical flow trajectories which adds to robustness; and (ii) CLNF also allows us to estimate the 3D head pose of the driver's face by minimizing  error in the projection of a generic 3D mesh model of the face w.r.t. the 2D location of landmarks in the image. The histogram features generated from the optical flow trajectories along with the 3D head pose features (yaw, pitch and row), give us $\phi(\text{face})\in\mathbb{R}^{12}$ when using the CLNF tracker. 

In Section~\ref{sec:experiments} we present results with the features from KLT, as well as the results with richer features obtained from the CLNF model.

\subsection{Outside-vehicle features.}\label{subsec:outside_features} 
The outside feature vector $\ve{x}_t$ encodes the information about the outside environment such as the road conditions, vehicle dynamics, etc. In order to get this information, we use the road-facing camera together with the vehicle's GPS coordinates, its speed, and the street maps. More specifically, we obtain two binary features from the road-facing camera   indicating whether a lane exists on the left side and on the right side of the  vehicle. We also augment the vehicle's GPS coordinates with the street maps and extract a binary feature indicating if the vehicle is within 15 meters of a road artifact  such as intersections, turns, highway exists, etc. We also encode the average,  maximum, and  minimum speeds of the vehicle over the last 5 seconds as features. This results in a $\ve{x}_t \in\mathbb{R}^6$ dimensional feature vector.

\section{Bayesian networks for\\ maneuver anticipation}
\label{sec:approach}

In this section we propose alternate Bayesian networks~\citep{Jain15} based on Hidden Markov Model (HMM) for maneuver anticipation. These models form a strong baseline to compare our sensory-fusion deep learning architecture. 

Driving maneuvers are influenced by multiple interactions involving the vehicle, its driver, outside traffic, and occasionally global factors like the driver's destination. These interactions influence the driver's intention, i.e. their state of mind before the maneuver, which is not directly observable.  In our Bayesian network formulation, we represent the driver's intention with discrete states that are  \textit{latent} (or hidden). In order to anticipate maneuvers, we  jointly model the driving context and the  \textit{latent} states in a tractable manner. We represent the driving context as a set of features described in Section~\ref{sec:features}. We now present the motivation for the Bayesian networks and then discuss our key model Autoregressive Input-Output HMM (AIO-HMM). 

\begin{figure}[t]
\centering
\includegraphics[width=.9\linewidth]{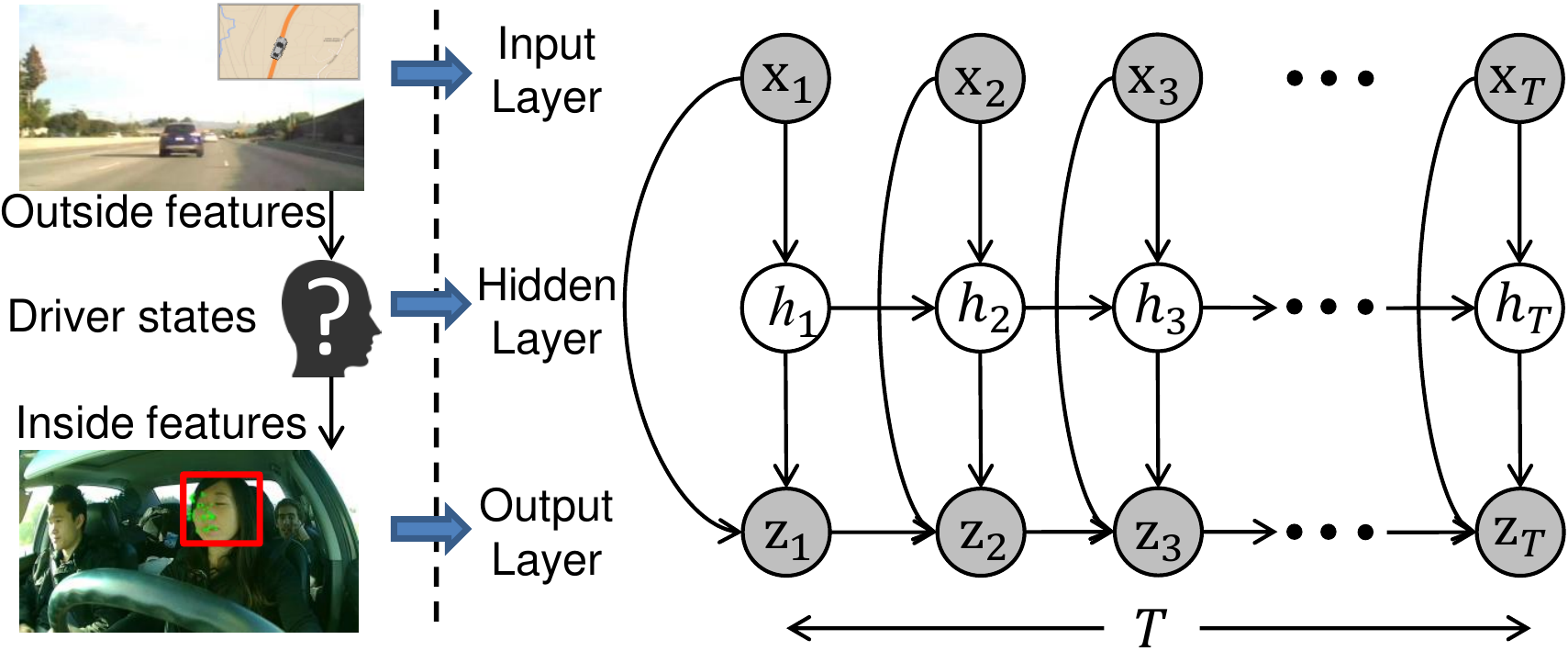}
\caption{\textbf{AIO-HMM.} The model has three layers: (i) Input (top): this  layer represents outside vehicle features $\ve{x}$; (ii) Hidden (middle): this  layer represents driver's latent states ${h}$; and (iii) Output (bottom): this  layer represents inside vehicle features $\ve{z}$. This layer also captures temporal dependencies of inside vehicle features. $T$ represents time.}
\label{fig:model}
\end{figure}

\subsection{Modeling driving maneuvers}
Modeling maneuvers require temporal modeling of the driving context. Discriminative methods, such as the Support Vector Machine and the Relevance Vector Machine~\citep{Tipping01}, which do not model the temporal aspect perform poorly on anticipation tasks, as we show in Section~\ref{sec:experiments}. Therefore, a temporal model such as the Hidden Markov Model (HMM) is better suited to model maneuver anticipation. 

An HMM models how the driver's \textit{latent} states generate both the inside driving context ($\ve{z}_t$) and the outside driving context ($\ve{x}_t$). However, a more accurate model should capture how events \textit{outside} the vehicle (i.e. the outside driving context) affect the driver's state of mind, which then generates the observations \textit{inside} the vehicle (i.e. the inside driving context). 
Such interactions can be modeled by an Input-Output HMM (IOHMM)~\citep{Bengio95}. However, modeling the problem with IOHMM does not capture the temporal dependencies of the inside driving context. These dependencies are critical to capture the smooth and temporally correlated behaviours such as the driver's face movements. We therefore present Autoregressive Input-Output HMM (AIO-HMM) which extends IOHMM to model these observation dependencies. Figure~\ref{fig:model} shows the AIO-HMM graphical model for modeling maneuvers. We learn separate AIO-HMM model for each maneuver. In order to anticipate maneuvers, during inference we determine which model best explains the past several seconds of the driving context based on the data log-likelihood.  In Appendix~\ref{sec:aiohmm} we describe the training and inference procedure for AIO-HMM.


\section{Experiments}
\label{sec:experiments}

\begin{figure}
\centering
\includegraphics[width=.9\linewidth]{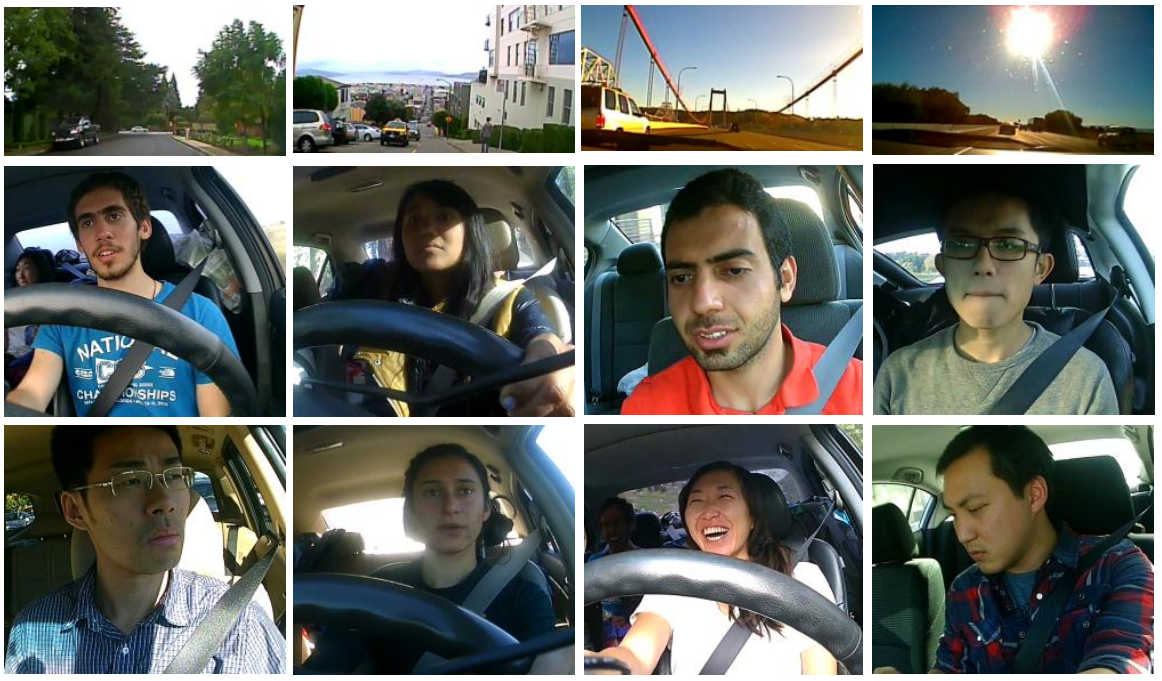}
\caption{\textbf{Our data set} is diverse in drivers and landscape.}
\label{fig:diverse_data}
\end{figure}

In this section we first give an overview of our data set and then present the quantitative results. We also demonstrate our system and algorithm on real-world driving scenarios. \textbf{Our video demonstrations are available at}: \hbox{\url{http://www.brain4cars.com}. }

\subsection{Driving data set}
\label{subsec:dataset}
Our data set consists of natural driving videos with both inside and outside views of the car, its speed, and the global position system (GPS) coordinates.\footnote{The inside and outside cameras operate at 25 and 30 frames/sec.} The outside car video captures the view of the road ahead. We collected this driving data set under fully natural settings without any intervention.\footnote{\textbf{Protocol:} We set up cameras, GPS and speed recording device in subject's personal vehicles and left it to record the data. The subjects were asked to ignore our setup and drive as they would normally.} 
It consists of 1180 miles of freeway and city driving and encloses 21,000 square miles across two states. We collected this data set from 10 drivers over a period of two months. The complete data set has a total of 2 million video frames and includes diverse landscapes. Figure~\ref{fig:diverse_data} shows a few samples from our data set. We annotated the driving videos with a total of 700 events containing 274 lane changes, 131 turns, and 295 randomly sampled instances of driving straight. Each lane change or turn annotation marks the start time of the maneuver, i.e., before the car touches the lane or yaws, respectively. For all annotated events, we also annotated the lane information, i.e., the number of lanes on the road and the current lane of the car. Our data set is publicly available at \hbox{\url{http://www.brain4cars.com}. }

\subsection{Baseline algorithms}
We compare the following algorithms:

\begin{itemize}
 \setlength{\itemsep}{3pt}
  \setlength{\parsep}{0pt}
\item \textit{Chance}: Uniformly randomly anticipates a maneuver.
\item \textit{SVM}~\citep{Morris11}: Support Vector Machine is a discriminative classifier~\citep{Cortes95}. Morris et al.~\citep{Morris11} takes this approach  for anticipating maneuvers.\footnote{Morries et al.~\citep{Morris11} considered binary 
classification problem (lane change vs driving straight) and used RVM~\citep{Tipping01}.}
We train the SVM on 5 seconds of driving context by concatenating all frame features 
to get a $\mathbb{R}^{3840}$ dimensional feature vector. 
\item \textit{Random-Forest}~\citep{Criminisi11}: This is also a discriminative classifier that
learns many decision trees from the training data, and at test time it
averages the prediction of the individual decision trees. We train
it on the same features as SVM with 150 trees of depth ten each.
\item \textit{HMM}: This is the Hidden Markov Model. 
We train the HMM on a temporal sequence of feature vectors that we extract every 0.8 seconds, 
i.e., every 20 video frames.  
We consider three versions of the HMM: (i) HMM $E$: with only outside features from the road camera, the vehicle's speed, GPS and street maps (Section~\ref{subsec:outside_features});
(ii) HMM $F$: with only inside features from the driver's face (Section~\ref{subsec:inside_features}); 
and (ii) HMM $E+F$: with both inside and outside features. 
\item \textit{IOHMM}: Jain et al.~\cite{Jain15} modeled driving maneuvers with this Bayesian network. It is trained on the same features as HMM $E + F$.
\item \textit{AIO-HMM}: Jain et al.~\citep{Jain15} proposed this Bayesian network for modeling maneuvers. It is trained on the same features as HMM $E + F$. 
\item \textit{Simple-RNN} (S-RNN): In this architecture sensor streams are fused by simple concatenation and then passed through a single RNN with LSTM units. 
\item \textit{Fusion-RNN-Uniform-Loss} (F-RNN-UL): In this architecture sensor streams are passed through separate RNNs, and the high-level representations from RNNs are then fused via a fully-connected layer. The loss at each time step takes the form $-\log(y_t^k)$. 
\item \textit{Fusion-RNN-Exp-Loss} (F-RNN-EL): This architecture is similar to F-RNN-UL, except that the loss exponentially grows with time $-e^{-(T-t)}\log(y_t^k)$. 
\end{itemize}

Our RNN and LSTM implementations are open-sourced and available at \texttt{NeuralModels}~\citep{Neuralmodels}. For the RNNs in our Fusion-RNN architecture we use a single layer LSTM of size 64 with sigmoid gate activations and tanh activation for hidden representation. Our fully connected fusion layer uses tanh activation and outputs a 64 dimensional vector. Our overall architecture (F-RNN-EL and F-RNN-UL) have nearly 25,000 parameters that are learned using RMSprop~\citep{Dauphin15}. 

\begin{algorithm}[t]\caption{Maneuver anticipation}
\begin{algorithmic}
\STATE \textbf{Initialize} $m^* = \textit{driving straight}$\\
\STATE \textbf{Input} Features $\{(\ve{x}_1,...,\ve{x}_T),(\ve{z}_1,...,\ve{z}_T)\}$ and prediction threshold $p_{th}$
\STATE \textbf{Output} Predicted maneuver $m^*$
\WHILE{$t=1$ to $T$}
        \STATE Observe features $(\ve{x}_1,...,\ve{x}_t)$ and $(\ve{z}_1,...,\ve{z}_t)$
        \STATE Estimate probability $ \ve{y}_t$ of each maneuver in $\mathcal{M}$
        \STATE $m_t^*=\argmax_{m\in\mathcal{M}}\ve{y}_t$
		\IF{$m_t^* \neq \textit{driving straight}$ \& $\ve{y}_t\{m_t^*\} > p_{th} $}        
			\STATE $m^* = m_t^*$
			\STATE \textbf{break}
		\ENDIF

\ENDWHILE
\STATE \textbf{Return} $m^*$
\end{algorithmic}
\label{alg:inference}
\end{algorithm}

\begin{table*}[t!]
\centering
\caption{{\textbf{Maneuver Anticipation Results.} Average \textit{precision}, \textit{recall} and \textit{time-to-maneuver} are computed from 5-fold cross-validation. Standard error is also shown. Algorithms are compared on the features from Jain et al.~\citep{Jain15}.}}
\resizebox{1\textwidth}{!}{
\centering
\begin{tabular}{cr|ccc|ccc|ccc}
&  &\multicolumn{3}{c}{Lane change}&\multicolumn{3}{|c}{Turns}&\multicolumn{3}{|c}{All maneuvers}\\
\cline{1-11}
\multicolumn{2}{c|}{\multirow{2}{*}{Method}} & \multirow{2}{*}{$Pr$ (\%)}  & \multirow{2}{*}{$Re$ (\%)} & Time-to-  & \multirow{2}{*}{$Pr$ (\%)} & \multirow{2}{*}{$Re$ (\%)} & Time-to-  & \multirow{2}{*}{$Pr$ (\%)} & \multirow{2}{*}{$Re$ (\%)}  & Time-to- \\ 
& & & &  maneuver (s) &  & &  maneuver (s) &  & & maneuver (s)\\\hline
&Chance	&	33.3		&	33.3		&	-	&	33.3		&	33.3		&	-	&	20.0		&	20.0		&	-\\
&Morris et al.~\citep{Morris11} SVM	&	73.7 $\pm$ 3.4	&	57.8 $\pm$ 2.8	&	2.40 		&	64.7 $\pm$ 6.5	&	47.2 $\pm$ 7.6	&	2.40 		&	43.7 $\pm$ 2.4	&	37.7 $\pm$ 1.8	& 1.20\\
&Random-Forest &   71.2 $\pm$ 2.4  &   53.4 $\pm$ 3.2  &   3.00    &   68.6 $\pm$ 3.5  & 44.4 $\pm$ 3.5  &   1.20      &   51.9 $\pm$ 1.6 &   27.7 $\pm$ 1.1  & 1.20\\
&HMM $E$ & 75.0 $\pm$ 2.2 & 60.4 $\pm$ 5.7 & 3.46  &  74.4 $\pm$ 0.5 & 66.6 $\pm$ 3.0 & 4.04  & 63.9 $\pm$ 2.6 & 60.2 $\pm$ 4.2 & 3.26 \\
&HMM $F$		&	76.4 $\pm$ 1.4	&	75.2 $\pm$ 1.6	&	3.62		&	75.6 $\pm$ 2.7	&		60.1 $\pm$ 1.7	&	3.58 		&	64.2 $\pm$ 1.5	&	36.8 $\pm$ 1.3	&	2.61\\
&HMM $E+F$	&	80.9 $\pm$ 0.9	&	79.6 $\pm$ 1.3	&	3.61 		&	73.5 $\pm$ 2.2	&
75.3 $\pm$ 3.1	&	4.53 		&	67.8 $\pm$ 2.0	&	67.7 $\pm$ 2.5	&	3.72 \\\hline
&IOHMM		&	81.6 $\pm$ 1.0	&	{79.6 $\pm$ 1.9}	&	3.98  	&	77.6 $\pm$ 3.3	&		{75.9 $\pm$ 2.5}	&	4.42 		&	74.2 $\pm$ 1.7	&	71.2 $\pm$ 1.6	&	3.83 \\
&(\textit{Our final Bayesian network}) AIO-HMM		&	{83.8 $\pm$ 1.3}	&	79.2 $\pm$ 2.9	&	3.80  		&	{80.8	 $\pm$ 3.4}	&		75.2 $\pm$ 2.4	&	4.16 		&	{77.4 $\pm$ 2.3}	&	{71.2 $\pm$ 1.3}	&	3.53 \\
& S-RNN & 85.4 $\pm$ 0.7 & 86.0 $\pm$ 1.4 & 3.53 & 75.2 $\pm$ 1.4 & 75.3 $\pm$ 2.1 & 3.68 & 78.0 $\pm$ 1.5 & 71.1 $\pm$ 1.0 & 3.15 \\
&F-RNN-UL & \textbf{92.7} $\pm$ 2.1 & 84.4 $\pm$ 2.8 & 3.46 & 81.2 $\pm$ 3.5 & 78.6 $\pm$ 2.8 & 3.94 & 82.2 $\pm$ 1.0 & 75.9 $\pm$ 1.5 & 3.75 \\
&(\textit{Our final deep architecture})  F-RNN-EL & 88.2 $\pm$ 1.4 & \textbf{86.0} $\pm$ 0.7 & 3.42 & \textbf{83.8} $\pm$ 2.1 & \textbf{79.9} $\pm$ 3.5 & 3.78 & \textbf{84.5} $\pm$ 1.0 & \textbf{77.1} $\pm$ 1.3 & 3.58\\
\end{tabular}
}
\label{tab:prscore}
\end{table*}

\subsection{Evaluation protocol}

We evaluate an algorithm based on its correctness in predicting future maneuvers. We anticipate maneuvers every 0.8 seconds where the algorithm processes the recent context and assigns a probability to each of the four maneuvers: \{\textit{left lane change, right lane change, left turn, right turn}\} and a probability to the event of \textit{driving straight}. These five probabilities together sum to one.
After anticipation, i.e. when the algorithm has computed all five probabilities, the algorithm predicts a  maneuver if its probability is above a threshold $p_{th}$. If none of the maneuvers' probabilities are above this threshold, the algorithm does not make a maneuver prediction and predicts \textit{driving  straight}. However, when it predicts one of the four maneuvers, it sticks with this prediction and makes no further predictions for next 5 seconds or until a maneuver occurs, whichever happens earlier. After 5 seconds or a maneuver has occurred, it returns to anticipating  future maneuvers. Algorithm~\ref{alg:inference} shows the inference steps for maneuver anticipation.

During this process of anticipation and prediction, the algorithm makes (i) true predictions ($tp$): when it predicts the correct maneuver; (ii) false predictions ($fp$): when it predicts a maneuver but the driver performs a different  maneuver; (iii) false positive predictions ($fpp$): when it predicts a maneuver but the driver does not perform any maneuver (i.e. \textit{driving straight}); and (iv) missed predictions ($mp$): when it predicts \textit{driving straight} but the driver performs a maneuver. We evaluate the algorithms using their precision and recall scores:
$$Pr = \frac{tp}{\underbrace{tp+fp+fpp}_\text{Total \# of maneuver predictions}};\;\;\;Re=\frac{tp}{\underbrace{tp+fp+mp}_\text{Total \# of maneuvers}}$$
The precision measures the fraction of the predicted maneuvers that are correct and recall measures the 
fraction of the maneuvers that are correctly predicted. For true predictions ($tp$) we also compute the 
average \textit{time-to-maneuver}, where time-to-maneuver is  the interval between the  time of algorithm's prediction and the start of the maneuver.

We perform cross validation to choose the number of the driver's latent states in the AIO-HMM and the threshold on  probabilities for maneuver prediction. For \textit{SVM} we cross-validate for the parameter $C$ and the choice of kernel from Gaussian and polynomial kernels. The parameters are chosen as the ones giving the highest F1-score on a validation set. 
The F1-score is the harmonic mean of the precision and recall, defined as $F1 = 2*Pr*Re/(Pr+Re)$.

\subsection{Quantitative results}

We evaluate the algorithms on maneuvers that were not seen during training  and report the results using 5-fold cross validation. Table \ref{tab:prscore} reports the precision and recall scores 
under three settings: (i) \textit{Lane change}: when the algorithms only predict for the left and right lane changes. This setting is relevant for highway driving where the prior probabilities of turns are low; (ii) \textit{Turns}: when the algorithms only predict for the left and right turns; and (iii) \textit{All maneuvers}: here the algorithms jointly predict all four maneuvers. All three settings include the instances of \textit{driving straight}. 

Table~\ref{tab:prscore} compares the performance of the baseline anticipation algorithms, Bayesian networks, and the variants of our deep learning model. All algorithms in Table~\ref{tab:prscore} use same feature vectors and KLT face tracker which ensures a fair comparison. As shown in the table, overall the best algorithm for maneuver anticipation is F-RNN-EL, and the best performing Bayesian network is AIO-HMM. F-RNN-EL significantly outperforms AIO-HMM in every setting. This improvement in performance is because RNNs with LSTM units are very expressive models with an internal memory. This allows them to model the much needed long temporal dependencies for anticipation. Additionally, unlike AIO-HMM,  F-RNN-EL is a discriminative model that does not make any assumptions about the generative nature of the problem.  The results also highlight the importance of modeling the temporal nature in the data. Classifiers like SVM and Random Forest do not model the temporal aspects and hence performs poorly. 

The performance of several variants of our deep architecture, reported in Table~\ref{tab:prscore},  justifies our design decisions to reach the final fusion architecture. When predicting all maneuvers, F-RNN-EL gives 6\%  higher precision and recall than S-RNN, which performs a simple fusion by concatenating the two sensor streams. On the other hand, F-RNN models each sensor stream with a separate RNN and then uses a fully connected layer to fuse the high-level representations at each time step. This form of sensory fusion is  more principled since the sensor streams represent different data modalities. 
In addition, exponentially growing the loss further improves the performance. Our new loss scheme  penalizes the network proportional to the length of context it has seen.  When predicting all maneuvers, we observe that F-RNN-EL shows an improvement of 2\% in precision and recall over F-RNN-UL. We conjecture that exponentially growing the loss acts like a regularizer. It reduces the risk of our network over-fitting early in time when there is not enough context available. Furthermore, the time-to-maneuver remains comparable for F-RNN  with and without exponential loss. 

The Bayesian networks AIO-HMM and HMM $E+F$ adopt different sensory fusion strategies. AIO-HMM fuses the two sensory streams using an input-output model, on the other hand HMM $E+F$ performs early fusion by concatenation. As a result, AIO-HMM gives 10\% higher precision than HMM $E+F$ for jointly predicting all the maneuvers. AIO-HMM further extends IOHMM by modeling the temporal dependencies of events inside the vehicle. This results in better performance: on average AIO-HMM precision is 3\% higher than IOHMM, as shown in Table~\ref{tab:prscore}. Another important aspect of anticipation is the joint modeling of the inside and outside driving contexts. HMM $F$ learns only from the inside driving context, while HMM $E$ learns only from the outside driving context. The performances of both the models is therefore less than HMM $E+F$, which learns jointly both the contexts. 

Table~\ref{tab:fpp} compares the $fpp$ of different algorithms.  False positive predictions ($fpp$) happen when an algorithm predicts a maneuver but the driver does not perform any maneuver (i.e. drives straight). Therefore low value of $fpp$ is preferred. HMM $F$ performs best on this metric at 11\%
as it mostly assigns a high probability to \textit{driving straight}. 
However, due to this reason, it incorrectly predicts \textit{driving straight} even when maneuvers happen. This results in the low recall of \hbox{HMM $F$} at 36\%, as shown in Table~\ref{tab:prscore}. AIO-HMM's 
$fpp$ is 10\% less than that of IOHMM and HMM $E+F$, and F-RNN-EL is 3\% less than AIO-HMM. The primary reason for false positive predictions is distracted driving. Drivers interactions with fellow passengers or their looking at the surrounding scenes are sometimes wrongly interpreted by the algorithms. Understanding driver distraction is still an open problem, and orthogonal to the objective of this work. 

\begin{table}[t]
\centering
\caption{\textbf{False positive prediction} ($fpp$) of different algorithms. The number inside parenthesis is the standard error.}
\resizebox{.85\linewidth}{!}{
\begin{tabular}{r|ccc}
Algorithm	&	Lane change	&	Turns	&	All\\\hline
Morris et al.~\citep{Morris11} SVM	&	15.3 (0.8) &		13.3 (5.6)	&	24.0 (3.5)\\
Random-Forest & 16.2 (3.3) & 12.9 (3.7) & 17.5 (4.0) \\
HMM $E$ & 36.2 (6.6) & 33.3 (0.0) & 63.8 (9.4) \\
HMM $F$	&	23.1 (2.1)	&	23.3 (3.1)	&	11.5 (0.1)\\
HMM $E+F$	&	30.0 (4.8)	&	21.2 (3.3)	&	40.7	 (4.9)\\
IOHMM	&	28.4 (1.5)	&	25.0 (0.1)	&	40.0 (1.5)\\
AIO-HMM	&	24.6 (1.5)	&	20.0 (2.0)	&	30.7 (3.4)\\
S-RNN & 16.2 (1.3) & 16.7 (0.0) & 19.2 (0.0)\\
F-RNN-UL & 19.2 (2.4) & 25.0 (2.4) & 21.5 (2.1) \\
F-RNN-EL & 10.8 (0.7) & 23.3 (1.5) & 27.7 (3.8)
\end{tabular}
}
\label{tab:fpp}
\end{table}

\begin{table}[h]
\centering
\caption{\textbf{3D head-pose features.} In this table we study the effect of better features with best performing algorithm from Table~\ref{tab:prscore} in `All maneuvers' setting. We use~\citep{Baltrusaitis13} to track 68 facial landmark points and estimate 3D head-pose.}
{
\newcolumntype{P}[2]{>{\footnotesize#1\hspace{0pt}\arraybackslash}p{#2}}
\setlength{\tabcolsep}{2pt}
\centering
\resizebox{\hsize}{!}{
\begin{tabular}
{@{}p{0.40\linewidth}| P{\centering}{16mm}P{\centering}{16mm}P{\centering}{16mm}@{}}
\multirow{2}{*}{Method} & \multirow{2}{*}{$Pr$ (\%)}  & \multirow{2}{*}{$Re$ (\%)} & Time-to-  \\ 
 & & &  maneuver (s)\\\hline
F-RNN-EL &  {84.5} $\pm$ 1.0 & {77.1} $\pm$ 1.3 & 3.58\\
F-RNN-EL w/ 3D head-pose &  \textbf{90.5} $\pm$ 1.0 & \textbf{87.4} $\pm$ 0.5 & 3.16\\
\end{tabular}
}}
\label{tab:features}
\end{table}

\begin{figure*}[t]
\centering
\begin{subfigure}[b]{.23\textwidth}
	\includegraphics[width=\linewidth]{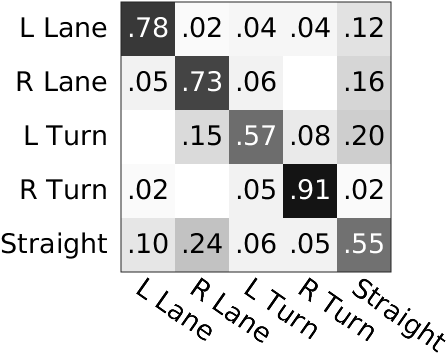}
	\caption{IOHMM}
	\end{subfigure}
\begin{subfigure}[b]{.23\textwidth}
	\includegraphics[width=\linewidth]{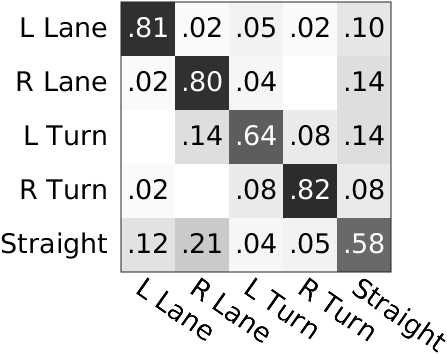}
	\caption{AIO-HMM}
	\end{subfigure}	
	\begin{subfigure}[b]{.23\textwidth}
	\includegraphics[width=\linewidth]{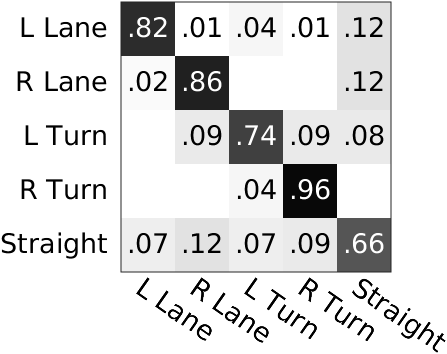}
	\caption{F-RNN-EL}
	\end{subfigure}	
	\begin{subfigure}[b]{.23\textwidth}
	\includegraphics[width=\linewidth]{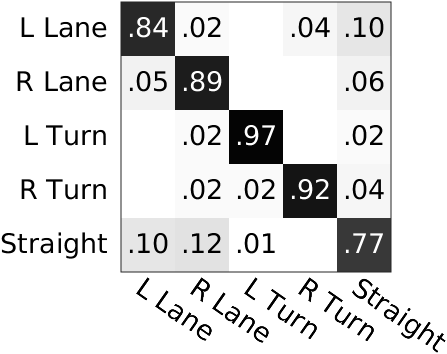}
	\caption{F-RNN-EL w/ 3D-pose}
	\end{subfigure}
	\caption{\textbf{Confusion matrix} of different algorithms when jointly predicting all the maneuvers. Predictions made by algorithms are represented by rows and actual maneuvers are represented by columns. Numbers on the diagonal represent precision.}
	\label{fig:confmat}	
\end{figure*}

\textbf{3D head-pose features.} The modularity of our approach allows experimenting with more advanced head tracking algorithms. We replace the pipeline for extracting features from the driver's face~\citep{Jain15} by a Constrained Local Neural Field (CLNF) model~\citep{Baltrusaitis13}. The new vision pipeline tracks 68 facial landmark points and estimates the driver's 3D head pose as described in Section~\ref{sec:features}. As shown in Table~\ref{tab:features}, we see a significant, 6\% increase in precision and 10\% increase in recall of F-RNN-EL when using features from our new vision pipeline. This increase in performance is attributed to the following reasons: (i) robustness of CLNF model to variations in illumination and head pose; (ii) 3D head-pose features are very informative for understanding the driver's intention; and (iii) optical flow trajectories generated by tracking facial landmark points represent head movements better, as shown in Figure~\ref{fig:feature_compare}. The confusion matrix in Figure~\ref{fig:confmat} shows the precision for each maneuver. F-RNN-EL gives a higher precision than AIO-HMM on every maneuver when both algorithms are trained on same features (Fig.~\ref{fig:confmat}c). The new vision pipeline with CLNF tracker further improves the precision of F-RNN-EL on all maneuvers (Fig.~\ref{fig:confmat}d). 

\begin{figure}[t]
\centering
\includegraphics[width=0.8\linewidth]{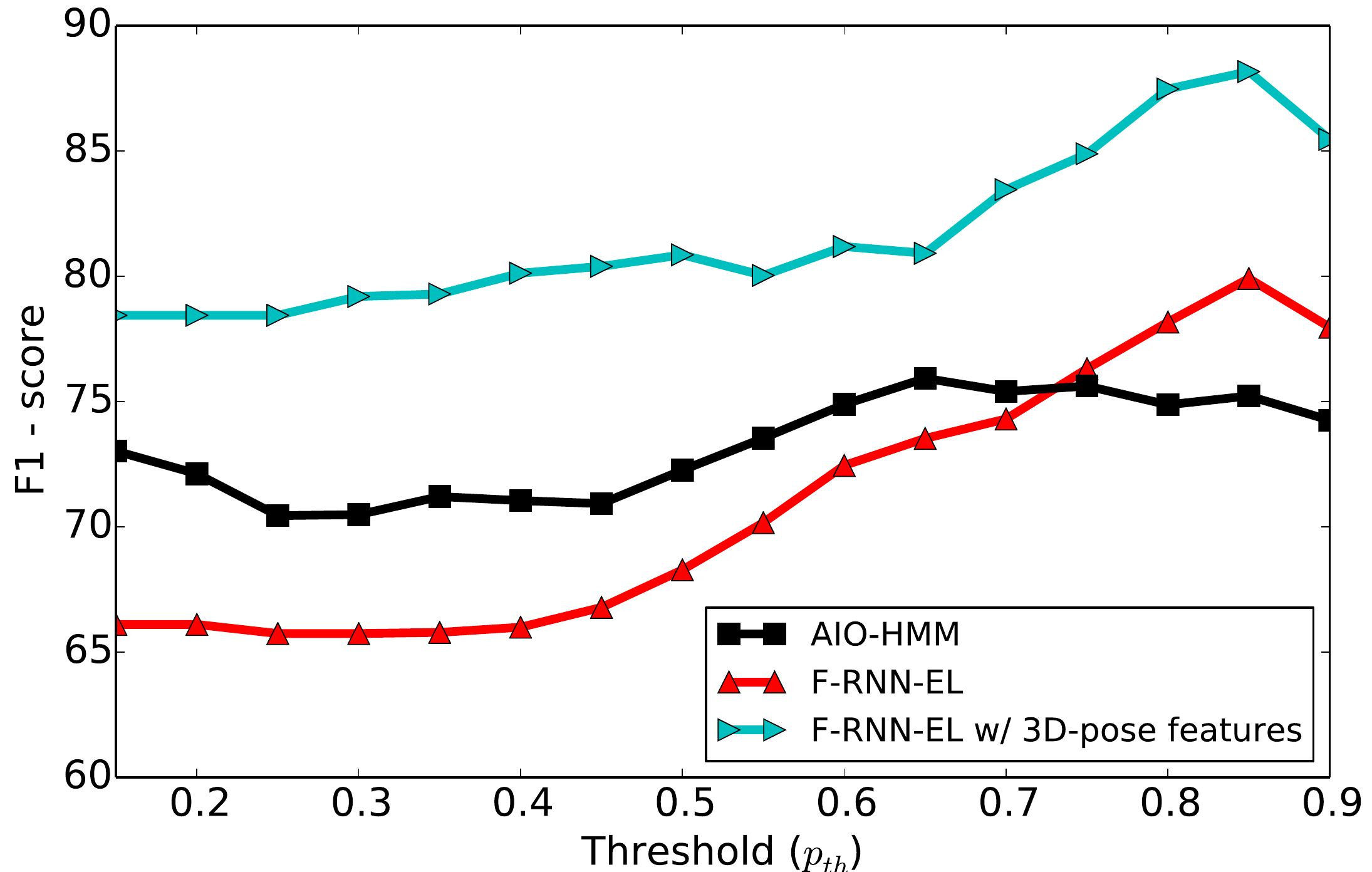}
\caption{\textbf{Effect of prediction threshold $p_{th}$.} At test time an algorithm makes a prediction only when it is at least $p_{th}$ confident in its prediction. This plot shows how F1-score vary with change in prediction threshold.}
\label{fig:f1score}
\end{figure}

\textbf{Effect of prediction threshold.} In Figure~\ref{fig:f1score} we study how F1-score varies as we change the prediction threshold $p_{th}$.  We make the following observations: (i) The F1-score does not undergo large variations with changes to the prediction threshold. Hence, it allows practitioners  to fairly trade-off between the precision and recall without hurting the F1-score by much; and (ii)   the maximum F1-score attained by F-RNN-EL is 4\% more than AIO-HMM when compared on the same features and 13\% more with our new vision pipeline.
In {Tables~\ref{tab:prscore},~\ref{tab:fpp} and \ref{tab:features}}, we used the threshold values which gave the highest F1-score.

\noindent\textbf{Anticipation complexity.} The F-RNN-EL anticipates maneuvers every 0.8 seconds using the previous 5 seconds of the driving context. The complexity mainly comprises of feature extraction and the model inference in Algorithm~\ref{alg:inference}. Fortunately both these steps can be performed as a dynamic program by storing the computation of the most recent anticipation. Therefore, for every anticipation we only process the incoming 0.8 seconds and not complete 5 seconds of the driving context. On average we predict a maneuver under 0.20~milliseconds using Theano~\citep{Bastien12} on Nvidia K40 GPU on Ubuntu~12.04. 

\section{Conclusion}

In this paper we considered the problem of anticipating driving maneuvers a few seconds before the driver performs them.  This problem requires the modeling of long temporal dependencies and the fusion of multiple sensory streams. We proposed a novel deep learning architecture based on Recurrent Neural Networks (RNNs) with Long Short-Term Memory (LSTM) units for anticipation. Our architecture learns to fuse multiple sensory streams, and by training it in a sequence-to-sequence prediction manner, it explicitly learns to anticipate using only a partial temporal context. We also proposed a novel loss layer for anticipation which prevents over-fitting. 

We release an open-source data set of 1180 miles of natural driving. We performed an extensive evaluation and showed improvement over many baseline algorithms. Our sensory fusion deep learning approach gives a precision of 84.5\% and recall of 77.1\%, and anticipates maneuvers \hbox{3.5 seconds} (on average) before they happen. By incorporating the driver's 3D head-pose our precision and recall improves to 90.5\% and 87.4\% respectively. Potential application of our work is enabling advanced driver assistance systems (ADAS) to alert drivers before they perform a dangerous maneuver, thereby giving drivers more time to react. We believe that our deep learning architecture is widely applicable to many activity anticipation problems. Our code and data set are publicly available on the project web-page. 
\\~\\~
\textbf{Acknowledgement.} We thank NVIDIA for the donation of K40 GPUs used in
this research. We also thank Silvio Savarese for useful discussions. This work
was supported by National Robotics Initiative (NRI) award 1426452, Office of Naval
Research (ONR) award N00014-14-1-0156, and by Microsoft Faculty Fellowship and NSF Career
Award to Saxena.

\appendices
\section{Modeling Maneuvers with AIO-HMM}
\label{sec:aiohmm}

Given $T$ seconds long driving context $\mathcal{C}$  before the maneuver $\mathcal{M}$, we learn a generative model for the context $P(\mathcal{C}|\mathcal{M})$. The driving context $\mathcal{C}$ consists of the outside driving context and the inside driving context. The outside and inside contexts are temporal sequences represented by the outside features $\ve{x}_1^T = \{\ve{x}_1,..,\ve{x}_T\}$ and the inside features $\ve{z}_1^T = \{\ve{z}_1,..,\ve{z}_T\}$ respectively. The corresponding sequence of the driver's latent states is ${h}_1^T = \{{h}_1,..,{h}_T\}$. $\ve{x}$ and $\ve{z}$ are vectors and ${h}$ is a discrete state.
\begin{align}
\nonumber P(\mathcal{C}|\mathcal{M}) &= \sum_{h_1^T} P(\ve{z}_1^T,\ve{x}_1^T,h_1^T|\mathcal{M})\\
\nonumber &= P(\ve{x}_1^T|\mathcal{M})\sum_{h_1^T} P(\ve{z}_1^T,h_1^T|\ve{x}_1^T,\mathcal{M})\\
\label{eq:aio-hmm} & \propto \sum_{h_1^T} P(\ve{z}_1^T,h_1^T|\ve{x}_1^T,\mathcal{M})
\end{align}
We model the correlations between $\ve{x}$, $h$ and $\ve{z}$ with an AIO-HMM as shown in Figure~\ref{fig:model}. The AIO-HMM models the distribution in equation~\eqref{eq:aio-hmm}. It does not assume any generative process for the outside features $P(\ve{x}_1^T|\mathcal{M})$. It instead models them in a discriminative manner. 
The top (input) layer of the AIO-HMM consists of  outside features $\ve{x}_1^T$. The outside features then affect the driver's latent states $h_1^T$, represented by the middle (hidden) layer, which then generates the inside features $\ve{z}_1^T$ at the bottom (output) layer. The events inside the vehicle such as the driver's head movements are temporally correlated because they are generally smooth. The AIO-HMM handles these dependencies with autoregressive connections in the output layer.  

\noindent \textbf{Model Parameters.}
AIO-HMM has two types of parameters: (i) state transition parameters $\mathbf{w}$; and (ii) observation emission parameters ($\boldsymbol\mu$,$\mathbf{\Sigma}$). We use set $\mathcal{S}$ to denote the possible latent states of the driver. For each state $h=i\in\mathcal{S}$, we parametrize transition probabilities of leaving the state with log-linear functions, and parametrize the output layer feature emissions with normal distributions. 
\begin{align*}
\text{Transition: }& P(h_t = j | h_{t-1} = i,
\ve{x}_t;\mathbf{w}_{ij}) = \frac{e^{\mathbf{w}_{ij} \cdot \ve{x}_t}}{\sum_{l \in \mathcal{S}}
e^{\mathbf{w}_{il} \cdot \ve{x}_t}}\\
\text{Emission: }&P(\ve{z}_t | h_t=i,\ve{x}_t,\ve{z}_{t-1};\boldsymbol\mu_{it},\mathbf{\Sigma}_i) =
\mathcal{N}(\ve{z}_t|\boldsymbol\mu_{it},\mathbf{\Sigma}_i) 
\end{align*}

The inside (vehicle) features represented by the output layer are jointly influenced by all three layers. These interactions are modeled by the mean and variance of the normal distribution. We model the mean of the distribution using the outside and inside features from the vehicle as follows:
$$\boldsymbol\mu_{it} = (1 + \mathbf{a}_i \cdot \ve{x}_t + \mathbf{b}_i \cdot \ve{z}_{t-1})
\boldsymbol\mu_i$$
In the equation above, $\mathbf{a}_i$ and $\mathbf{b}_i$ are parameters that we learn for every state $i \in \mathcal{S}$. Therefore, the parameters we learn for state $i \in \mathcal{S}$ are $\boldsymbol\theta_i = \{\boldsymbol\mu_i$, $\mathbf{a}_i$, $\mathbf{b}_i$, $\mathbf{\Sigma}_i$ and $\mathbf{w}_{ij} | j \in \mathcal{S}\}$, and the overall model parameters are $\boldsymbol\Theta = \{\boldsymbol\theta_i|i\in\mathcal{S}\}$.

\subsection{Learning AIO-HMM parameters}

The training data $\mathcal{D} = \{(\ve{x}_{1,n}^{T_n},\ve{z}_{1,n}^{T_n})|n=1,..,N\}$ consists of $N$ instances of a maneuver $\mathcal{M}$. The goal is to maximize the data log-likelihood.
\begin{equation}
\label{eq:likelihood}
l(\boldsymbol\Theta;\mathcal{D}) = \sum_{n=1}^N
\log P(\ve{z}_{1,n}^{T_n}|\ve{x}_{1,n}^{T_n};\boldsymbol\Theta)
 \end{equation}
Directly optimizing  equation~\eqref{eq:likelihood}  is challenging because parameters $h$ representing the driver's states are \textit{latent}. We therefore use the iterative EM procedure to learn the model parameters. In EM, instead of directly maximizing equation~\eqref{eq:likelihood}, we maximize its simpler lower bound. We estimate the lower bound in the E-step and then maximize that estimate in the M-step. These two steps are repeated iteratively. 

\noindent \textbf{E-step.} In the E-step we get the lower bound of equation~\eqref{eq:likelihood} by calculating the expected
value of the \textit{complete} data log-likelihood using the current estimate of the parameter $\hat{\boldsymbol\Theta}$. 
\begin{equation}
\label{eq:estep}
\text{E-step: }Q(\boldsymbol\Theta;\hat{\boldsymbol\Theta}) =
E[l_c(\boldsymbol\Theta;\mathcal{D}_c)|\hat{\boldsymbol\Theta},\mathcal{D}] 
\end{equation}
where $l_c(\boldsymbol\Theta;\mathcal{D}_c)$ is the  log-likelihood 
 of  the \textit{complete} data $\mathcal{D}_c$ defined as:  
\begin{align}
\mathcal{D}_c &= \{(\ve{x}_{1,n}^{T_n},\ve{z}_{1,n}^{T_n},h_{1,n}^{T_n})|n=1,..,N\} \label{eq:complete-data}\\
l_c(\boldsymbol\Theta;\mathcal{D}_c) &= \sum_{n=1}^N
\log P(\ve{z}_{1,n}^{T_n},h_{1,n}^{T_n}|\ve{x}_{1,n}^{T_n};\boldsymbol\Theta) \label{eq:complete-likelihood}
\end{align}

We should note that the occurrences of hidden variables $h$ in $l_c(\boldsymbol\Theta;\mathcal{D}_c)$ are marginalized in equation~\eqref{eq:estep}, and hence $h$ need not be known.
We efficiently estimate $Q(\boldsymbol\Theta;\hat{\boldsymbol\Theta})$ using the forward-backward algorithm~\citep{Murphy12}. 

\noindent \textbf{M-step.} In the M-step we maximize the expected value  of the complete data log-likelihood $Q(\boldsymbol\Theta;\hat{\boldsymbol\Theta})$ and update the model parameter as follows:
\begin{equation}
\label{eq:mstep}
\text{M-step: }\boldsymbol\Theta = \argmax_{\boldsymbol\Theta}
Q(\boldsymbol\Theta;\boldsymbol\hat{\boldsymbol\Theta})
\end{equation}

Solving  equation~\eqref{eq:mstep} requires us to optimize for the parameters 
$\boldsymbol\mu$, $\mathbf{a}$, $\mathbf{b}$, $\mathbf{\Sigma}$ and $\mathbf{w}$. We optimize all parameters expect $\mathbf{w}$ exactly by deriving their closed form update expressions. We optimize $\mathbf{w}$ using the gradient descent. 

\subsection{Inference of Maneuvers}

Our learning algorithm trains separate AIO-HMM models for each maneuver. The goal during inference is to determine which model best explains the past $T$ seconds of the driving context not seen during training. We evaluate the likelihood of the inside and outside feature sequences ($\ve{z}_1^T$ and $\ve{x}_1^T$)  for each maneuver, and anticipate the probability $P_\mathcal{M}$ of each maneuver $\mathcal{M}$ as follows:
\begin{align}
\label{eq:infer} P_\mathcal{M} &= P(\mathcal{M}|\ve{z}_1^T, \ve{x}_1^T) \propto P(\ve{z}_1^T, \ve{x}_1^T | \mathcal{M})P(\mathcal{M})
\end{align}
Algorithm~\ref{alg:coactive} shows the complete inference procedure. The inference in equation~\eqref{eq:infer} simply requires a forward-pass~\citep{Murphy12} of the AIO-HMM, the complexity of which is 
\hbox{$\mathcal{O}(T(|\mathcal{S}|^2 + |\mathcal{S}||\ve{z}|^3 + |\mathcal{S}||\ve{x}|))$}. However, in practice it is only $\mathcal{O}(T|\mathcal{S}||\ve{z}|^3)$ because $|\ve{z}|^3 \gg |S|$ and $|\ve{z}|^3 \gg |\ve{x}|$. Here $|\mathcal{S}|$
is the number of discrete states representing the driver's intention, while $|\ve{z}|$ and $|\ve{x}|$ are the dimensions of the inside and outside feature vectors respectively. In equation~\eqref{eq:infer} $P(\mathcal{M})$ is the prior probability of maneuver $\mathcal{M}$. We assume an uninformative uniform prior over the maneuvers.
  \begin{algorithm}[H]\caption{Anticipating maneuvers }
\begin{algorithmic}
\INPUT Driving videos, GPS, Maps and Vehicle Dynamics
\OUTPUT Probability of each maneuver
\STATE Initialize the face tracker with the driver's face
\WHILE{$driving$}
	\STATE Track the driver's face~\citep{Viola04}
	\STATE Extract features $\ve{z}_1^T$ and $\ve{x}_1^T$ (Sec.~\ref{sec:features})
	\STATE Inference $P_\mathcal{M} = P(\mathcal{M}|\ve{z}_1^T, \ve{x}_1^T)$ (Eq.~\eqref{eq:infer})
	\STATE Send the inferred probability of each maneuver to ADAS
\ENDWHILE
\end{algorithmic}
\label{alg:coactive}
\end{algorithm}

{
\bibliographystyle{plainnat}
\bibliography{longstrings,references}
}

\end{document}